\makeatletter

\PassOptionsToPackage{svgnames}{xcolor}

\PassOptionsToPackage{noend}{algorithmic}
\PassOptionsToPackage{noend}{algpseudocode}

\AddToHook{package/lineno/before}{\RequirePackage{amsmath}}

\AtBeginDocument{
  \@ifpackageloaded{theapa}{
    \NewDocumentCommand{\citet}{o m}{
      \IfNoValueTF{#1}
        {\citeauthor{#2} (\citeyear{#2})}
        {\citeauthor{#2} (\citeyear[#1]{#2})}
    }
    \NewDocumentCommand{\citep}{o m}{
      \IfNoValueTF{#1}
        {\cite{#2}}
        {\cite[#1]{#2}}
    }
  }{
  }
}

\@ifpackageloaded{algorithmic}{
  \renewcommand{\algorithmiccomment}[1]{\hfill #1}
  \def\State\STATE
  \def\If\IF
  \def\Then\THEN
  \def\Elsif\ELSIF
  \def\Else\ELSE
  \def\Endif\ENDIF
  \def\For\FOR
  \def\Forall\FORALL
  \def\Do\DO
  \def\Endfor\ENDFOR
  \def\While\WHILE
  \def\Endwhile\ENDWHILE
  \def\Repeat\REPEAT
  \def\Until\UNTIL
  \def\Return\RETURN
  \def\Require\REQUIRE
  \def\Ensure\ENSURE
  \def\Comment\COMMENT
}{}

\@ifpackageloaded{algpseudocode}{
  \algrenewcommand{\algorithmiccomment}[1]{\hfill #1}
  \algblockdefx[method]{Method}{EndMethod}[2]{\textbf{method} \textsc{#1} (#2)}{}
  \algblockdefx[function]{Method}{EndMethod}[2]{\textbf{method} \textsc{#1} (#2)}{}
  \def\STATE\State
  \def\IF\If
  \def\THEN\Then
  \def\ELSIF\ElsIf
  \def\ELSE\Else
  \def\ENDIF\EndIf
  \def\FOR\For
  \def\FORALL\ForAll
  \def\DO\Do
  \def\ENDFOR\EndFor
  \def\WHILE\While
  \def\ENDWHILE\EndWhile
  \def\REPEAT\Repeat
  \def\UNTIL\Until
  \def\RETURN\Return
  \def\REQUIRE\Require
  \def\ENSURE\Ensure
  \def\COMMENT\Comment
}{}

\makeatother

\documentclass{article}
\usepackage{ijcai24}
\usepackage{times}
\usepackage{soul}
\usepackage[hidelinks]{hyperref}
\usepackage[utf8]{inputenc}
\usepackage[small]{caption}
\usepackage[switch]{lineno}
\makeatletter
\@ifpackageloaded{theapa}{
}{
 \usepackage{natbib}
}
\makeatother

\usepackage{xparse}

\usepackage{amsmath}
\usepackage{amsthm}
\usepackage{amssymb}
\usepackage{amsfonts}

\usepackage{tabularx}
\usepackage{booktabs}   
\usepackage{multirow}
\usepackage{subfigure}

\usepackage{manfnt}             

\usepackage{xspace}             
\usepackage{relsize}            
\usepackage{adjustbox}          
\usepackage{diagbox}            
\usepackage{pdflscape}          

\usepackage{microtype}

\usepackage{graphicx}
\usepackage{url}
\usepackage{comment}            
\usepackage{xcolor}             
\usepackage{colortbl}

\usepackage{algorithm}
\usepackage{algorithmic} 

\usepackage{xr}

\usepackage{bm}
\usepackage{alphabeta}          
\usepackage{stmaryrd}

\def\eqref#1{equation~\ref{#1}}

\def\1{\bm{1}}
\def\0{\bm{0}}

\def\rh{{\textnormal{h}}}
\def\ri{{\textnormal{i}}}

\def\rs{{\textnormal{s}}}

\def\rx{{\textnormal{x}}}

\DeclareMathAlphabet{\mathsfit}{\encodingdefault}{\sfdefault}{m}{sl}
\SetMathAlphabet{\mathsfit}{bold}{\encodingdefault}{\sfdefault}{bx}{n}

\def\E{{\mathbb{E}}}

\def\N{{\mathcal{N}}}

\def\R{{\mathbb{R}}}

\def\T{{\mathcal{T}}}

\def\X{{\mathcal{X}}}

\def\Z{{\mathbb{Z}}}

\DeclareMathOperator*{\argmax}{arg\,max}
\DeclareMathOperator*{\argmin}{arg\,min}

\newcommand{\brackets}[1]{{\left<#1\right>}}
\newcommand{\braces}[1]{{\left\{#1\right\}}}
\newcommand{\parens}[1]{{\left(#1\right)}}

\NewDocumentCommand{\diffby}{s m O{}}{
 \IfBooleanTF{#1}
  {\frac{\partial#3}{\partial#2}}
  {\frac{d#3}{d#2}}
}

\newcommand{\satisfies}{\vDash}

\RenewDocumentCommand{\to}{o o}{
 \IfNoValueTF{#1}
  {\rightarrow}
  {\IfNoValueTF{#2}
   {\xrightarrow{#1}}
   {\xrightarrow[#2]{#1}}}
}

\NewDocumentCommand{\affect}{o o}{
 \IfNoValueTF{#1}
  {\rightsquigarrow}
  {
   \IfNoValueTF{#2}
   {\rightsquigarrow^{#1}}
   {\rightsquigarrow^{#1}_{#2}}
  }
}

\newcommand{\erf}{\function{erf}}
\newcommand{\erfc}{\function{erfc}}

\newcommand{\erfcx}{\function{erfcx}}

\makeatletter

\usepackage{pdftexcmds}
\usepackage{etoolbox}
\newtoggle{dev}
\togglefalse{dev}

\ifcase\pdf@shellescape
  \or
  \toggletrue{dev}
  \or
\fi

\usepackage{stackengine}
\stackMath
\newcommand\tsup[2][2]{
 \def\useanchorwidth{T}
  \ifnum#1>1
    \stackon[-.5pt]{\tsup[\numexpr#1-1\relax]{#2}}{\scriptscriptstyle\sim}
  \else
    \stackon[.5pt]{#2}{\scriptscriptstyle\sim}
  \fi
}

\newcommand{\pddl}[1]{\textsf{\small #1}}

\newcommand{\function}[1]{\textsc{#1}}

\def\_{\\[-0.3em]}

\newlength{\maxwidth}

\let\@myref\ref

\newcommand{\refsec}[1]{Sec.\,\@myref{#1}}
\newcommand{\refseq}[1]{Sec.\,\@myref{#1}}
\newcommand{\refig}[1]{Fig.\,\@myref{#1}}
\newcommand{\reftbl}[1]{Table \@myref{#1}}
\newcommand{\refstep}[1]{Step \@myref{#1}}
\newcommand{\refalgo}[1]{Alg.\,\@myref{#1}}
\newcommand{\refchap}[1]{Chap.\,\@myref{#1}}
\newcommand{\reflst}[1]{List \@myref{#1}}
\newcommand{\refeq}[1]{Eq.\,\@myref{#1}} 

\newcommand{\reftheo}[1]{Thm.\,\@myref{#1}}
\newcommand{\refline}[1]{line\,\@myref{#1}}
\newcommand{\refdef}[1]{Def.\, \@myref{#1}}
\newcommand{\refex}[1]{Example\,\@myref{#1}}
\newcommand{\refconv}[1]{Conv.\,\@myref{#1}}
\newcommand{\reffact}[1]{Fact.\,\@myref{#1}}
\newcommand{\reflemma}[1]{Lemma.\,\@myref{#1}}
\newcommand{\refcorol}[1]{Col.\,\@myref{#1}}

\newcommand{\refsecs}[2]{Sec.\,\@myref{#1}-\@myref{#2}}
\newcommand{\refseqs}[2]{Sec.\,\@myref{#1}-\@myref{#2}}
\newcommand{\refigs}[2]{Fig.\,\@myref{#1}-\@myref{#2}}
\newcommand{\reftbls}[2]{Tables \@myref{#1}-\@myref{#2}}
\newcommand{\refsteps}[2]{Steps \@myref{#1}-\@myref{#2}}
\newcommand{\refalgos}[2]{Alg.\,\@myref{#1}-\@myref{#2}}
\newcommand{\refchaps}[2]{Chap.\,\@myref{#1}-\@myref{#2}}
\newcommand{\reflsts}[2]{Lists \@myref{#1}-\@myref{#2}}
\newcommand{\refeqs}[2]{Eq.\,\@myref{#1}-\@myref{#2}}
\newcommand{\refpages}[2]{p.\pageref{#1}-\@myref{#2}}
\newcommand{\reftheos}[2]{Thm.\,\@myref{#1}-\@myref{#2}}
\newcommand{\reflines}[2]{line\,\@myref{#1}-\@myref{#2}}
\newcommand{\refdefs}[2]{Def.\,\@myref{#1}-\@myref{#2}}
\newcommand{\refexs}[2]{Example\,\@myref{#1}-\@myref{#2}}
\newcommand{\refconvs}[2]{Conv.\,\@myref{#1}-\@myref{#2}}
\newcommand{\reffacts}[2]{Facts.\,\@myref{#1}-\@myref{#2}}
\newcommand{\reflemmas}[2]{Lemma.\,\@myref{#1}-\@myref{#2}}
\newcommand{\refcorols}[2]{Col.\,\@myref{#1}-\@myref{#2}}

\newcounter{list}[section]

\makeatother

\makeatletter

\newcommand{\pre}{\function{pre}}

\newcommand{\adde}{\function{add}}
\newcommand{\dele}{\function{del}}
\newcommand{\cost}{\function{cost}}

\def\hash{\text{\relsize{-1}\#}}
\newcommand{\ar}[1]{\hash{}#1}

\newcommand{\sota}{State-of-the-Art\xspace}

\newcommand{\astar}{\ifmmode{A^*}\else{A$^*$}\fi\xspace}
\newcommand{\gbfs}{\ifmmode{\mathrm{GBFS}}\else{GBFS}\fi\xspace}
\NewDocumentCommand{\uct}{s}{\ifmmode{\mathrm{UCT}{\IfBooleanT{#1}{^*}}}\else{UCT{\IfBooleanT{#1}{*}}}\fi\xspace}
\NewDocumentCommand{\guct}{s}{\ifmmode{\mathrm{GUCT}{\IfBooleanT{#1}{^*}}}\else{GUCT{\IfBooleanT{#1}{*}}}\fi\xspace}

\newcommand{\newheuristic}[2]{
 \def#1{
  \relax\ifmmode
  h^\mathrm{#2}\xspace
  \else
  \text{#2}\xspace
  \fi
 }
}

\newheuristic{\lmcut}{LMcut}
\newheuristic{\mands}{M\&S}
\newheuristic{\pdb}{PDB}
\newheuristic{\ff}{FF}
\newheuristic{\ce}{CEA}
\newheuristic{\cg}{CG}
\newheuristic{\gc}{GC}
\newheuristic{\ad}{add}
\newheuristic{\hmax}{max}
\newheuristic{\lc}{LC}
\newheuristic{\blind}{blind}

\newcommand{\newlearnedheuristic}[2]{
 \def#1{
  \relax\ifmmode
  H^\mathrm{#2}\xspace
  \else
  \text{#2}\xspace
  \fi
 }
}

\newlearnedheuristic{\Hlmcut}{LMcut}
\newlearnedheuristic{\Hmands}{M\&S}
\newlearnedheuristic{\Hpdb}{PDB}
\newlearnedheuristic{\Hff}{FF}
\newlearnedheuristic{\Hce}{CEA}
\newlearnedheuristic{\Hcg}{CG}
\newlearnedheuristic{\Had}{add}
\newlearnedheuristic{\Hmax}{max}
\newlearnedheuristic{\Hlc}{LC}
\newlearnedheuristic{\Hblind}{blind}

\newcommand{\newUnitCostHeuristic}[2]{
 \def#1{
  \relax\ifmmode
  \hat{h}^\mathrm{#2}\xspace
  \else
  \text{#2}\xspace
  \fi
 }
}

\newUnitCostHeuristic{\lmcuto}{LMcut}
\newUnitCostHeuristic{\mandso}{M\&S}
\newUnitCostHeuristic{\ffo}{FF}
\newUnitCostHeuristic{\ceo}{CEA}
\newUnitCostHeuristic{\cgo}{CG}
\newUnitCostHeuristic{\ado}{add}
\newUnitCostHeuristic{\gco}{GoalCount}
\newUnitCostHeuristic{\lco}{LC}

\newcommand{\newrandomheuristic}[2]{
 \def#1{
  \ifmmode
  \rh^\mathrm{#2}\xspace
  \else
  \text{#2}\xspace
  \fi
 }
}

\newrandomheuristic{\rlmcut}{LMcut}
\newrandomheuristic{\rmands}{M\&S}
\newrandomheuristic{\rpdb}{PDB}
\newrandomheuristic{\rff}{FF}
\newrandomheuristic{\rce}{CEA}
\newrandomheuristic{\rcg}{CG}
\newrandomheuristic{\rad}{add}
\newrandomheuristic{\rhmax}{max}
\newrandomheuristic{\rlc}{LC}

\makeatother

\usepackage{xkeyval}
\makeatletter

\define@key{strips}{negative-preconditions}[1]{\def\@conditiontype{#1}}
\define@key{strips}{disjunctive-preconditions}[2]{\def\@conditiontype{#1}}
\define@key{strips}{conditional-effects}[1]{\def\@usecondeffect{#1}}
\define@key{strips}{action-costs}[1]{\def\@usecost{#1}}
\define@key{strips}{derived-predicates}[1]{\def\@useaxiom{#1}}
\define@key{strips}{lifted}[1]{\def\@lifted{#1}}
\define@key{strips}{optimal}[1]{\def\@optimal{#1}}
\define@key{strips}{unitcost}[1]{\def\@unitcost{#1}}

\def\conditionset{
\if\@useaxiom0
P
\else
P\cup P_X
\fi
}

\let\satisfies@orig\satisfies

\def\satisfies{
\if\@conditiontype0
\supseteq
\else
\satisfies@orig
\fi
}

\def\condition{
\if\@conditiontype0
\conditionset
\else
\mathcal{F}(\conditionset)
\fi
}

\def\ga{
\if\@lifted0
a
\else
a^{\dagger}
\fi
}

\def\applyformula{
\if\@usecondeffect0
(s \setminus \dele(a)) \cup \adde(a)
\else
(s
 \setminus \braces{e \mid (c \triangleright e) \in \dele(\ga), c\satisfies s})
 \cup      \braces{e \mid (c \triangleright e) \in \adde(\ga), c\satisfies s}
\fi
}

\NewDocumentCommand{\strips}{O{}}{
\def\@useaxiom{0}
\def\@conditiontype{0}
\def\@usecondeffect{0}
\def\@usecost{0}
\def\@lifted{0}
\def\@optimal{0}
\def\@unitcost{0}
\setkeys{strips}{#1}
\if\@lifted1
  \strips@propositional\par
  \strips@lifted
\else
  \strips@propositional
\fi
}

\newcommand{\strips@propositional}{
\if\@conditiontype1
Given a set of propositional variables $V$,
let $\mathcal{F}(V)$ be a propositional formula consisting of $V$ and
logical operations $\braces{\land,\lnot}$.
\fi
\if\@conditiontype2
Given a set of propositional variables $V$,
let $\mathcal{F}(V)$ be a propositional formula consisting of $V$ and
logical operations $\braces{\land,\lor,\lnot}$.
\fi
\if\@useaxiom0
We define a propositional STRIPS Planning problem
as a 4-tuple $\brackets{P,A,I,G}$
where
 $P$ is a set of propositional variables,
 $A$ is a set of actions,
 $I\subseteq P$ is the initial state, and
 $G\subseteq \conditionset$ is a goal condition.
\else
We define a propositional STRIPS Planning problem
as a 6-tuple $\brackets{P,A,X,P_X,I,G}$
where
 $P$ is a set of propositions,
 $A$ is a set of actions,
 $X$ is a set of axioms,
 $P_X$ is a set of derived propositions ($P\cap P_X=\emptyset$),
 $I\subset P$ is the initial state, and
 $G\subset \conditionset$ is a goal condition.
\fi
\if\@usecost0
Each action $a\in A$ is a 3-tuple $\brackets{\pre(a),\adde(a),\dele(a)}$ where
\else
Each action $a\in A$ is a 4-tuple $\brackets{\pre(a),\adde(a),\dele(a),\cost(a)}$ where
$\cost(a) \in \Z^{0+}$ is a cost,
\fi
$\pre(a) \subseteq \condition$ is a precondition and
\if\@usecondeffect0
$\adde(a), \dele(a)\subseteq P$ are the add-effects and delete-effects, respectively.
\else
$\adde(a), \dele(a)$ are the add-effects and delete-effects.
Each effect is denoted as $c \triangleright e$ where
$c \in \condition$ is an \emph{effect condition} and
$e \in P$.
\fi
\if\@useaxiom1
The set of axioms $X$ consists of clauses $f \Rightarrow p$ where
$f \in \condition$ is a body and $p \in P_X$ is a head.
\fi
A state $s\subseteq \conditionset$ is a set of true propositions
(all of $P\setminus s$ are false),
an action $a$ is \emph{applicable} when $s \satisfies \pre(a)$ (read: $s$ \emph{satisfies} $\pre(a)$),
and applying action $a$ to $s$ yields a new successor state
\if\@useaxiom0
$a(s) = \applyformula$.
\else
$a(s)$.
To compute $a(s)$, we first obtain a non-derived state
$a'(s) = \applyformula \setminus P_X $.
Then we perform a fix-point calculation such that
$s \gets s \cup \braces{p \in P_X \mid (f\Rightarrow p)\in X \land s \satisfies f}$
where $s$ is initialized to $a'(s)$.
\fi

The task of classical planning is to find a sequence of actions called a \emph{plan} $(\ga_1,\cdots,\ga_n)$
where, for $1\leq t\leq n$,
 $s_0=I$, $s_t\satisfies \pre(a_{t+1})$, $s_{t+1}=a_{t+1}(s_t)$,
 and $s_n\satisfies G$.
\if\@optimal1
 A plan is \emph{optimal} if
 \if\@usecost0
   there is no shorter plan.
 \else
   there is no plan with lower \emph{cost-to-go} $\sum_t \cost(a_t)$.
 \fi
 A plan is otherwise called \emph{satisficing}.
 \if\@usecost1
  \if\@unitcost1
  In this paper, we assume unit-cost: $\forall a\in A; \cost(a)=1$.
  \fi
 \fi
\fi
}

\newcommand{\strips@lifted}{
In a \emph{Lifted STRIPS} problem $\brackets{P,O,A,I,G}$,
each propositional variable is an \emph{instantiation}/\emph{grounding} of
a first-order logic predicate $P$.
Each predicate $p(x_1,\ldots,x_{\ar{p}})$ is parameterized by a list of parameters/variables/arguments $(x_1,\ldots,x_{\ar{p}})$,
where $\ar{p}$ is an \emph{arity} of $p$.
A proposition is obtained by substituting each $x_i$ with an \emph{object} in a set $O$.
Each $p$ therefore has $O^{\ar{p}}$ instantiations.
Similarly, each action is now called a \emph{ground action},
which is an instantiation of a \emph{lifted action} $a(x_1,\ldots,x_{\ar{p}})\in A$
parameterized by $\ar{a}$ parameters.
A ground action is obtained by substituting the arguments as well as
the parameters used in the preconditions and the effects.
}

\makeatother

\iftoggle{dev}{
  \IfFileExists{\jobname.needpyg}{
    \usepackage[finalizecache,cachedir=.]{minted}
  }{
    \usepackage{minted}
  }
}{
  \usepackage[frozencache,cachedir=.]{minted}
}

\usepackage{wrapfig}

\allowdisplaybreaks

\usepackage{listings}

\lstdefinelanguage{PDDL}
{
  breaklines=true,
  sensitive=false,    
  morecomment=[l]{;}, 
  alsoletter={:,-},   
  morekeywords={
    define,domain,problem,not,and,or,when,forall,exists,either,
    :domain,:requirements,:types,:objects,:constants,
    :predicates,:action,:parameters,:precondition,:effect,
    :fluents,:primary-effect,:side-effect,:init,:goal,
    :strips,:adl,:equality,:typing,:conditional-effects,
    :negative-preconditions,:disjunctive-preconditions,
    :existential-preconditions,:universal-preconditions,:quantified-preconditions,
    :functions,assign,increase,decrease,scale-up,scale-down,
    :metric,minimize,maximize,
    :durative-actions,:duration-inequalities,:continuous-effects,
    :durative-action,:duration,:condition
  }
}

\title{On Using Admissible Bounds for Learning Forward Search Heuristics}

\author{
Carlos Núñez-Molina$^1$\and Masataro Asai$^2$\and
Pablo Mesejo$^1$\And Juan Fernández-Olivares$^1$\\
\affiliations
$^1$University of Granada, Spain\\
$^2$MIT-IBM Watson AI Lab, USA\\
\emails
ccaarlos@ugr.es, masataro.asai@ibm.com, pmesejo@ugr.es, faro@decsai.ugr.es
}

\begin{document}
\maketitle

\begin{abstract}
In recent years, there has been growing interest in utilizing modern machine learning techniques to learn heuristic functions for forward search algorithms.
Despite this, there has been little theoretical understanding of
\emph{what} they should learn, \emph{how} to train them, and \emph{why} we do so.
This lack of understanding has resulted in the adoption of diverse training targets
(suboptimal vs optimal costs vs admissible heuristics)
and loss functions (e.g., square vs absolute errors) in the literature.
In this work, we focus on how to effectively utilize the information provided by admissible heuristics in heuristic learning.
We argue that learning from poly-time admissible heuristics
by minimizing mean square errors (MSE) is not the correct approach,
since its result is merely a noisy, inadmissible copy
of an efficiently computable heuristic.
Instead, we propose to model the learned heuristic as a \textit{truncated} gaussian,
where admissible heuristics are used not as training targets but as lower bounds of this distribution.
This results in a different loss function from the MSE commonly employed in the literature,
which implicitly models the learned heuristic as a gaussian distribution.
We conduct experiments where both MSE and our novel loss function are applied to learning a heuristic from optimal plan costs.
Results show that our proposed method
converges faster during training and yields better heuristics.
\end{abstract}

\section{Introduction}

\begin{figure}[t]
 \centering
 \includegraphics[width=.95\linewidth]{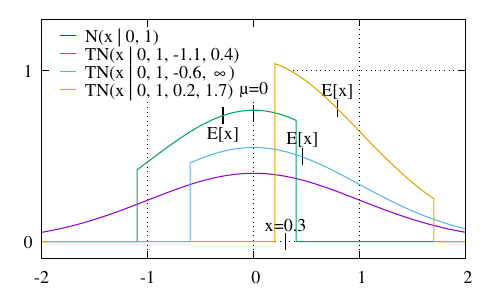}
 \caption{
 The probability density functions (PDFs) of Truncated Gaussian distributions $p(\rx)=\T\N(\mu=0,\sigma=1,l,u)$ with several lower/upper bounds $(l,u)$.
 In the heuristic learning setting,
 $\rx$ is the optimal solution cost $h^*$ sampled from the dataset and
 $\mu=\mu_{\theta}(s)$ is the prediction associated with a state $s$.
 The $(l,u)=(0.2,1.7)$ variant (yellow) shows that
 the mean $\E_{p(\rx)}[\rx]$,
 which we use as the search heuristic,
 respects the bounds $(l,u)$
 even when the predicted $\mu=0$ lies outside $(l,u)$.
 }
 \label{fig:truncated-gaussian}
\end{figure}

\begin{figure}[t]
    \includegraphics[width=\linewidth]{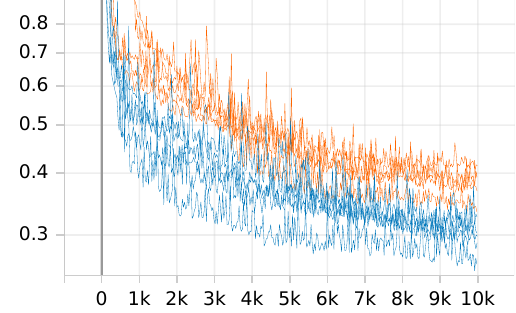}
    \caption{
    Comparison of the training curve ($x$-axis: training step) for the validation MSE loss ($y$-axis, logarithmic)
    between Gaussian (orange) and Truncated Gaussian (blue) models,
    independent runs recorded on 5 random seeds each.
    The losses converge faster for the latter due to the additional information
    provided by the admissible lower bound $l=\lmcut$.
    }
    \label{fig:loss-curve}
\end{figure}

Motivated by the success of Machine Learning (ML) approaches
in various decision making tasks \citep{dqn,alphago},
an increasing number of papers are tackling the problem of
learning a heuristic function for forward state space search in recent years.
Despite this interest, there has been little theoretical understanding of
\textit{what} these systems should learn, \textit{how} to train them and
\textit{why} we do so.
As a result, heuristic learning literature has adopted many different training targets
(corresponding to either admissible heuristics \citep{shen2020learning},
suboptimal solution costs \citep{ArfaeeZH11,ferber2022neural,marom2020utilising}
or optimal solution costs \citep{ernandes2004likely,shen2020learning})
and training losses (e.g.,
square errors \citep{shen2020learning},
absolute errors \citep{ernandes2004likely}
and piecewise absolute errors \citep{takahashi2019learning}).

In this work, we try to answer these questions from a statistical lens,
focusing on how to effectively utilize admissible heuristics in the context of heuristic learning.
We argue that learning from poly-time admissible heuristics, such as $\lmcut$ \citep{Helmert2009},
by minimizing mean square errors (MSE) does not provide any practical benefits,
since its result is merely a noisy, inadmissible copy of a heuristic that is already efficient to compute.
Then, if admissible heuristics should not be used as training targets,
how can we leverage them? In order to answer this question,
we first analyze the statistical implications behind the commonly used loss function, the MSE,
which implicitly models the learned heuristic as a Gaussian distribution.
Nonetheless, we contend that a better modeling choice for heuristics is given
by the \textit{Truncated} Gaussian distribution (\refig{fig:truncated-gaussian}),
due to the existence of \textit{bounds} on the values a heuristic can take
(e.g., heuristics never take on negative values).

The main contribution of this paper is a theoretically-motivated, statistical method for
learning an inadmissible heuristic while exploiting an admissible heuristic.
We propose to model the learned heuristic as a Truncated Gaussian,
where an admissible heuristic provides the lower bound of this distribution,
thus constraining heuristic predictions.
This modeling choice results in a loss function to be minimized that is different from the standard MSE loss.
We conduct extensive experimentation where
both loss functions are applied to learning heuristics from optimal plan costs in several classical planning domains.
Results show that those methods which model the learned heuristic as a Truncated Gaussian,
i.e., which are trained with our novel loss function,
learn faster and result in better heuristics
than those which model it as an ordinary Gaussian,
i.e., which are trained with the standard MSE loss.
To the best of our knowledge,
this is the first work that proposes the use of admissible heuristics
to constrain heuristic predictions and improve learning.\footnote{
The camera-ready version of this paper has been accepted for publication at IJCAI 2024.
Our full code and data can be found in github.com/pddl-heuristic-learning/pddlsl.
}

\section{Backgrounds}

\subsection{Classical Planning and Heuristics}
\strips[optimal,action-costs,unitcost]

A domain-independent heuristic function $h$ in classical planning is
a function of a state $s$ and the problem $\brackets{P,A,I,G}$.
It returns an estimate of the shortest (optimal) path cost from $s$ to one of the goal states (states that satisfy $G$),
typically through a symbolic, non-statistical means such as \textit{delete-relaxation},
a technique that ignores the delete-effects of actions in order to efficiently estimate the cost from $s$ to $G$.
The optimal cost-to-go, or a \emph{perfect heuristic}, is denoted by $h^*$.
A heuristic is called \textit{admissible} if it never overestimates it, i.e.,
$\forall s; 0\leq h(s)\leq h^*(s)$,
and \textit{inadmissible} otherwise.
Notable admissible heuristics include $\lmcut, \hmax$ and $h^+$ \citep{Helmert2009,bonet2001planning,hplus},
whereas $\ff$, $\ad$ and $\gc$ \citep{hoffmann01,bonet2001planning,FikesHN72} are prominent examples of inadmissible heuristics.

\subsection{Task: Supervised Learning for Heuristics}
\label{sec:supervised-heuristic-learning}

Let $p^*(\rx)$ be the unknown ground-truth probability distribution of (an) observable random variable(s) $\rx$
and let $p(\rx)$ be our current estimate of it.
Given a dataset $\X=\braces{x^{(1)},\ldots,x^{(N)}}$ of $N$ data points,
we denote an empirical data distribution as $q(\rx)$, which draws samples from $\X$ uniformly.
While often $q(\rx)$ may also be informally called a ground-truth distribution,
$q(\rx)$ is entirely different from either $p(\rx)$ or $p^*(\rx)$ because it is a distribution over a finite set of points,
i.e., a uniform mixture of dirac's delta $\delta$ distributions (\refeq{eq:empirical}).
Our goal is to obtain an estimate $p(\rx)$ that resembles $p^*(\rx)$ as closely as possible.
To do so, under the \emph{Maximum Likelihood Estimation} (MLE) framework,
we maximize the expectation of $p(\rx)$ over $q(\rx)$.
In other words, MLE tries to maximize the expected probability $p(x)$ of observing each data point $x \sim q(\rx)$:
\begin{align}
 q(\rx) &= \sum_\ri q(\rx|\ri)q(\ri)=\sum_{i=1}^N \delta(\rx=x^{(i)}) \cdot \frac{1}{N} \label{eq:empirical} \\
 p^*(\rx) &= \argmax_p \E_{q(\rx)} p(\rx)
 \notag \\
 &= \argmin_p \E_{q(\rx)} -\log p(\rx)
 \label{eq:mle}
\end{align}
Typically,
we assume $p^*(\rx)$ and $p(\rx)$ are of the same family of functions
parameterized by $\theta$, such as a set of neural network weights or the trees in random forests,
i.e., $p^*(\rx)=p_{\theta^*}(\rx)$, $p(\rx)=p_{\theta}(\rx)$.
This makes MLE a problem of finding the $\theta$ maximizing $\E_{q(\rx)}p_{\theta}(\rx)$.
Also, we typically minimize a \emph{loss} such as the \emph{negative log likelihood} (NLL) $-\log p(\rx)$,
since $\log$ is monotonic and preserves the optima $\theta^*$ (\refeq{eq:mle}).
Furthermore, $\E_{q(\rx)} \ldots$ is often estimated by Monte-Carlo sampling, e.g.,
$\E_{q(\rx)} -\log p(\rx) \approx \frac{1}{N} \sum_{i=1}^N -\log p(x_i)$,
where each $x_i$ is sampled from $q(\rx)$.

We further assume $p(\rx)$ to follow a specific distribution such as
a Gaussian distribution $\N(\mu,\sigma)$:
\begin{align}
 p(\rx)=\N(\rx\mid\mu,\sigma)=\frac{1}{\sqrt{2\pi\sigma^2}}e^{-\frac{(\rx-\mu)^2}{2\sigma^2}}.
\end{align}
We emphasize that
\textbf{the choice of the distribution determines the loss}.
When the model designer assumes $p(\rx)=\N(\mu,\sigma)$,
then the NLL is a shifted and scaled squared error:
\begin{align}
 \textstyle -\log p(\rx)=\frac{(\rx-\mu)^2}{2\sigma^2}+\log \sqrt{2\pi\sigma^2}.\label{eq:gaussian-nll}
\end{align}
Likewise, a Laplace distribution $L(\rx|\mu,b)=\frac{1}{2b}e^{-\frac{|\rx-\mu|}{b}}$
represents the absolute error because
its NLL is $\frac{|\rx-\mu|}{b} +\log 2b$.

The NLL loss is thus more fundamental and theoretically grounded than losses such as the Mean Squared Error,
although it is ``more complicated'' due to the division $\frac{1}{2\sigma^2}$ and the second term.
A reader unfamiliar with statistics may rightfully question
why such complications are necessary or
why $\sigma$ is not commonly used by the existing literature.
It is because many applications happen to require only a single prediction for a single input (\emph{point estimate}):
When we model the output distribution as a Gaussian $\N(\mu,\sigma)$,
we often predict $\mu$, which is simultaneously the mean and the mode of the distribution
and does not depend on $\sigma$.

Moreover, the MSE is a special case of the NLL that can be derived from it.
To derive the MSE,
we first simplify the loss into the squared error $(x-\mu)^2$
by setting $\sigma$ to an arbitrary constant, such as $\sigma=\frac{1}{\sqrt{2}}$,
because the variance/spread of the prediction does not matter in a point estimation of $\mu$.
As a result, we can also ignore the second term which is now a constant.
We then compute
the expectation $\E_{q(\rx)} (\rx-\mu)^2$ with a Monte-Carlo estimate
that samples $N$ data points $x_1,\ldots x_N \sim q(\rx)$,
predict $\mu=\mu_\theta(x_i)$ for each $x_i$ using a machine learning model $\mu_\theta$,
and compute the average: $\frac{1}{N}\sum_{i=1}^{N} (x_i-\mu_\theta(x_i))^2$.
In other words, \textbf{the MSE loss is nothing more than the Monte-Carlo estimate of the NLL loss of
a Gaussian with a fixed $\sigma=\frac{1}{\sqrt{2}}$}.
In contrast,
\textit{distributional estimates} represent the entire $p(\rx)$;
e.g., if $p(\rx)=\N(\mu,\sigma)$, then the model predicts both $\mu$ and $\sigma$.

The MLE framework can be applied to the supervised heuristic learning setting as follows.
Let $q(\rs,\rx)$ be the empirical data distribution,
where $\rs$ is a random variable representing a state-goal pair
(from now on, we will implicitly assume that states $\rs$ also contain goal information)
and $\rx$ a random variable representing the cost-to-go
(regardless of whether it corresponds to a heuristic estimate, optimal or suboptimal cost).
Then, the goal is to learn $p^*(\rx \mid \rs)$ where:
\begin{align}
  p^*(\rx \mid \rs) &= \argmax_\theta \E_{q(\rs,\rx)} p_\theta(\rx|\rs), \\
 p_\theta(\rx\mid \rs)
 &\textstyle=\N(\rx\mid \mu=\mu_{\theta}(\rs),\sigma=\frac{1}{\sqrt{2}})\label{eq:h-to-mu},
\end{align}
and $\mu_{\theta}(\rs)$ is the main body of the learned model,
such as a neural network parameterized by the weights $\theta$.
Supervised heuristic learning with distributional estimates is formalized similarly,
where the only difference is that an additional model (e.g. a neural network) with parameters $\theta_2$ predicts $\sigma$:
\begin{align}
 p_{(\theta_1,\theta_2)}(\rx\mid \rs)
 &\textstyle=\N(\rx\mid \mu=\mu_{\theta_1}(\rs),\sigma=\sigma_{\theta_2}(\rs)).\label{eq:h-to-mu-sigma}
\end{align}

\subsection{The Principle of Maximum Entropy}
The discussion above models $p(\rx)$ as a Gaussian distribution.
While the assumption of normality (i.e., following a Gaussian) is ubiquitous,
one must be able to justify such an assumption.
The \emph{principle of maximum entropy} \citep{maxent} states that $p(\rx)$
should be modeled as the maximum entropy (\textit{max-ent}) distribution
among all those that satisfy our constraints or assumptions,
where the entropy is defined as $\E_{p(x)}\brackets{-\log p(x)}$.
A set of constraints defines its corresponding max-ent distribution which,
being the \textit{most random} among those that satisfy those constraints,
minimizes assumptions other than those associated with the given constraints.
Conversely, a non max-ent distribution implicitly encodes additional or different assumptions
that can result in an accidental, potentially harmful bias.
For example, if we believe that our random variable $\rx$ has
a finite mean, a finite variance and a support/domain/range equal to $\R$,
then it \emph{must} be modeled as a Gaussian distribution according to this principle
because it is the max-ent distribution among all those that satisfy
these three constraints.

In other words,
a person designing a loss function of a machine learning model
must devise a reasonable set of constraints on the target variable $\rx$
to
identify the max-ent distribution $p(\rx)$ of the constraints,
which \emph{automatically} determines the \emph{correct} NLL loss for the model.
This paper tries to follow this principle as faithfully as possible.

\section{Utilizing Bounds for Learning}
\label{sec:training}

In the previous section, we provided some statistical background on heuristic learning. We now leverage this background to analyze
many of the decisions taken in the existing literature, sometimes unknowingly,
putting particular focus on how admissible heuristics are used during training.
Based on this analysis, we argue that the proper way of utilizing the information provided by admissible heuristics
is using them as the lower bound of a Truncated Gaussian distribution representing the learned heuristic.

We previously explained that
the heuristic to be learned is modeled as a probability distribution (e.g., a Gaussian),
instead of a single value:
The ML model is unsure about the true heuristic value $h^*$ associated with a state $s$.
When it predicts $\mu$, it believes not only that $\mu$ is the most likely value (the \emph{mode}) for $h^*$,
but also that other values are still possible.
The uncertainty of this prediction is given by $\sigma$:
The larger this parameter is, the more unsure the model is about its prediction.
The commonly used MSE loss is derived from the ad-hoc assumption that
$\sigma$ is fixed, i.e., independent from $s$,
which means that the model is equally certain (or uncertain) about $h^*$ for every state $s$.
This is unrealistic in most scenarios:
it is generally more difficult to accurately predict $h^*$ for
states that are further from the goal,
for which the uncertainty should be larger.
Therefore, the model should predict $\sigma$ in addition to $\mu$,
i.e., it should output a distributional estimate of $h^*$ instead of a point estimate.

Another crucial decision involves selecting \emph{what} to learn, i.e., the target / ground truth to use for the training.
It is easy to see that
training a model on a dataset
containing a practical (i.e., computable in polynomial time) heuristic,
admissible or otherwise, such as $\lmcut$ or $\ff$,
does not provide any practical benefits because,
even if the training is successful,
all we get is a noisy, lossy, slow copy of a heuristic that is already efficient to compute.
Worse, trained models always lose admissibility if the target is admissible.
To outperform existing poly-time heuristics,
i.e., achieve a \textit{super-symbolic benefit} from learning,
it is imperative to train the model on data of a better quality,
such as $h^+$ as proposed in \citet{shen2020learning}
or optimal solution costs $h^*$.
Although obtaining these datasets may prove computationally expensive in practice,
e.g., $h^+$ is NP-hard,
we can aspire to learn a heuristic that
outperforms the poly-time heuristics by training on these targets.

If poly-time admissible heuristics are not ideal training targets,
are they completely useless for learning a heuristic?
Intuitively this should not be the case, given the huge success of heuristic search
where they provide a strong search guidance toward the goal.
Our main question is then \emph{how} we should exploit the information they provide.
To answer this question, we must revise the assumption we previously made by using squared errors:
That $\rx=h^*$ follows a Gaussian distribution $\N(\mu,\sigma)$.
The issue with this assumption is that
$\N(\mu,\sigma)$ assigns a non-zero probability $p(x)$ to every $x \in \R$,
but we actually know that $h^*$ cannot take some values:
Given some admissible heuristic like $\lmcut$,
we know that $\lmcut \leq h^*$ holds for every state; therefore $p(x)=0$ when $x<\lmcut$.
Analogously, if for some state $s$
we know the cost $h^{sat}$ of a satisficing (non-optimal) plan from $s$ to
the goal, then $h^{sat}$ acts as an \textit{upper bound} of $h^*$.

According to the principle of maximum entropy, which serves as our \emph{why},
if we have a lower $l$ and upper $u$ bound for $h^*$,
then we should model $h^*$ using the max-ent
distribution with finite mean, finite variance, and a support equal to $(l,u)$,
which is
the \textit{Truncated Gaussian} distribution $\T\N(\rx | \mu,\sigma,l,u)$
as proven by \citet{dowson1973maximum},
formalized as \refeq{eq:truncated-gaussian}:
\begin{align}
 &\T\N(\rx | \mu,\sigma,l,u) =
 \left\{
 \begin{array}{cc}
  \frac{1}{\sigma}
   \frac{\phi(\frac{\rx-\mu}{\sigma})}
        {\Phi(\frac{u-\mu}{\sigma})-\Phi(\frac{l-\mu}{\sigma})} & l \leq \rx \leq u\\
  0 & \text{otherwise,}
 \end{array}
 \right.
 \label{eq:truncated-gaussian}
 \\
 &
 \text{where}\
 \textstyle
 \phi(\rx) = \frac{1}{\sqrt{2\pi}} \exp \frac{\rx^2}{2},\
 \Phi(\rx) = \frac{1}{2} (1 + \erf(\rx)),
 \notag
\end{align}
$l$ is the lower bound, $u$ is the upper bound,
$\mu$ is the pre-truncation mean, $\sigma$ is the pre-truncation standard deviation,
and $\erf$ is the error function.
$\T\N$ has the following NLL loss:
\begin{align}
 -\log \T\N(\rx | \mu,\sigma,l,u)
 &=\frac{(\rx-\mu)^2}{2\sigma^2} + \log \sqrt{2\pi\sigma^2} \label{eq:truncated-gaussian-nll} \\
 &+\textstyle\log \parens{\Phi\parens{\frac{u-\mu}{\sigma}}-\Phi\parens{\frac{l-\mu}{\sigma}}}.
 \notag
\end{align}

Modeling $h^*$ as a $\T\N$ instead of $\N$ presents several advantages.
Firstly, $\T\N$ constrains heuristic predictions
to lie in the range $(l,u)$ given by the bounds of the distribution.
Secondly, $\T\N$ generalizes $\N$ as $\T\N(\mu,\sigma,-\infty,\infty)=\N(\mu,\sigma)$ when no bounds are provided.
Finally, $\T\N$ opens the possibility for a variety of training scenarios for heuristic learning,
with a sensible interpretation of each type of data, including the satisficing solution costs.

In this work, we focus on the scenario where an admissible heuristic $h$ is
provided along with the optimal solution cost $h^*$ for each state, leaving other settings for future work.
In this case, $h$ acts as the lower bound $l$ of $h^*$,
which is modeled as a $\T\N(\rx=h^* | \mu,\sigma,h,\infty)$,
where $\mu$ and $\sigma$ are predicted by an ML model.
Note that we cannot use $h^*$ as $\T\N(h^* | \mu,\sigma,h^*,h^*)$
since, during evaluation/test time, we do not have access to the optimal cost $h^*$.
Also, this modeling decision is feasible even when no admissible heuristic is available
(e.g., when the PDDL description of the environment is not known, as in Atari games \citep{bellemare2013arcade})
since we can always resort to the blind heuristic $\blind(s)$ or simply do $l=0$,
which still results in a tighter bound than the one provided by
an untruncated Gaussian $\N(\mu,\sigma) = \T\N(\mu,\sigma,-\infty,\infty)$.

Finally, our setting is orthogonal and compatible with \textit{residual learning} \citep{yoon2008learning},
where the ML model does not directly predict $\mu$ but rather a
\textit{residual} or offset $\Delta\mu$ over a heuristic $h$, where $\mu=h + \Delta\mu$.
Residual learning can be seen as initializing the model output $\mu$
around $h$ which, when $h$ is a good \textit{unbiased estimator} of $h^*$,
facilitates learning.
This technique can be used regardless of whether $h^*$ is modeled as a
$\T\N$ or $\N$ because it merely corresponds to a particular implementation of $\mu=\mu_{\theta}(s)$,
which is used by both distributions.
Residual learning is analogous to the data normalization
commonly applied in standard regression tasks,
where features are rescaled and shifted to have mean 0 and variance 1.
However, residual learning is superior in the heuristic learning setting because
target data (e.g., $h^*$) is skewed above 0 and because the heuristic used as the basis for the residual
can handle out-of-distribution data due to its symbolic nature.

\subsection{Planning with a Truncated Gaussian}
\label{sec:testing}

At planning time,
we must obtain a point estimate of the output distribution,
which will be used as a heuristic to determine the ordering between search nodes.
As a point estimate, we can use any statistic of central tendency, thus we choose the mean.
It is important to note that
the $\mu$ parameter of $\T\N(\mu,\sigma,l,u)$ is \emph{not} the mean of this distribution
since $\mu$ corresponds to the mean of $\N(\mu,\sigma)$
(i.e., the mean of the distribution \textit{before truncation})
and does not necessarily lie in the interval $(l,u)$.
The mean of a Truncated Gaussian is obtained according to \refeq{eq:tg-mean}.
Note that a naive implementation of this formula results in rounding errors
(See the Appendix for a numerically stable implementation).
\begin{align}
 \E[\rx] = \mu +  \sigma\frac{\phi(\frac{l-\mu}{\sigma})-\phi(\frac{u-\mu}{\sigma})}{\Phi(\frac{u-\mu}{\sigma})-\Phi(\frac{l-\mu}{\sigma})}
 \label{eq:tg-mean}
\end{align}

\refeq{eq:tg-mean} satisfies $l\leq\E[\rx]\leq u$.
This means that, when a lower bound $l$ is provided (e.g., by an admissible heuristic),
the heuristic prediction returned by the model will never be smaller than $l$.
Analogously, when an upper bound $u$ is also provided (e.g., by a satisficing solution cost),
the model will never predict a heuristic value larger than $u$.
With this, we hope that
the use of a $\T\N$ during planning helps the model make predictions that are closer to $h^*$
than the bounds themselves,
potentially helping it achieve a super-symbolic improvement over admissible heuristics.

In contrast, the mode $\argmax_x p(x)$ of $\T\N$ is uninteresting:
While we could use it as another point estimate,
it is the same as the untruncated mean $\mu$ when the predicted $\mu$ is within the bounds,
and equal to one of the upper/lower bounds otherwise (see \refig{fig:truncated-gaussian}).
However, this inspires a naive alternative that is applicable even to $\N$,
which is to clip the heuristic prediction $\E[\rx]$ (equal to $\mu$ for $\N$) to the interval $[l,u]$.
We expect only a marginal gain from this trick because
it only improves \textit{really bad} predictions, i.e., those which would lie outside $[l,u]$ otherwise, and
does not affect predictions that correctly lie inside $[l,u]$.
In our experiments, we show that this approach is inferior to our first method.

We re-emphasize that despite the use of admissible heuristics during training the learned heuristic is inadmissible,
just like any learning-based heuristics proposed so far.
In case a distributional estimate is used, i.e., when the ML model also learns to predict $\sigma$,
we could discuss \textit{likely-admissibility} \citep{ernandes2004likely,marom2020utilising}.
However, this extension is left for future work.

\section{Experimental Evaluation}
\label{sec:sl-eval}

We evaluate the effectiveness of our new loss function
under the domain-specific generalization setting, where the learned heuristic function is required to generalize across
different problems of a single domain.
Due to space limitations,
we focus on the high-level descriptions and
describe the detailed configurations in the Appendix.

\paragraph{Data Generation.}

We trained our system on four classical planning domains:
\pddl{blocksworld-4ops}, \pddl{ferry}, \pddl{gripper}, and \pddl{visitall}.
Using PDDL domains as benchmarks for evaluating planning performance is a standard practice,
as exemplified by the International Planning Competitions (IPCs) \citep{vallati20152014}.
For each domain,
we generated three sets of problem instances (train, validation, test) with parameterized generators used in the IPCs.
We provided between 456 and 1536 instances for training
(the variation is due to the difference in the number of generator parameters in each domain),
between 132 and 384 instances for validation and testing (as separate sets),
and 100 instances sampled from the test set for planning.
The Appendix describes the domains and generator parameters.
Notably, the test instances are generated with larger parameters
in order to assess the generalization capability.
To generate the dataset from these instances,
we optimally solved each instance with \astar \citep{hart1968formal} and $\lmcut$ in Fast Downward \citep{Helmert2006}
under 5min runtime / 8GB memory (train,val) and 30min runtime / 8GB memory (test).
Whenever it failed to solve a instance within the limits,
we retried generation with a different random seed for a maximum of 20 times until success,
thus ensuring a specified number of instances were generated.
We also discarded trivial instances that satisfied the goal conditions at the initial state.
For each state $s$ in the optimal plan,
we archived $h^*$ and the values of several heuristics (e.g., $\lmcut$ and $\ff$).
Therefore, each instance was used to obtain several data points.

\paragraph{Model Configurations.}

We evaluated three different ML methods to show that our statistical model is implementation-agnostic.
Neural Logic Machine (NLM) \citep{dong2018nlm} is
an architecture designed for inductive learning and reasoning over symbolic data
which has been successfully applied to classical planning domains
for learning heuristic functions \citep{gehring2021pddlrl} with Reinforcement Learning \citep{sutton2018reinforcement}.
STRIPS-HGN \citep[HGN for short]{shen2020learning} is
another architecture based on the notion of \emph{hypergraphs}.
Lastly, we used linear regression with the hand-crafted features proposed by \citet{gomoluch2017towards},
which comprise the values of the goal-count \citep{FikesHN72} and FF \citep{hoffmann01} heuristics,
along with the total and mean number of effects ignored by FF's relaxed plan.

We analyze our learning \& planning system from several orthogonal axes.
\textbf{Gaussian vs.\ Truncated:}
Using $\mu(s)$ as the parameter of a Gaussian $\N(\mu(s), \sigma(s))$
or Truncated Gaussian $\T\N(\mu(s), \sigma(s), l, \infty)$ distribution.
\textbf{Learned vs.\ fixed sigma:}
Predicting $\sigma(s)$
or using a constant value $\sigma(s)=\frac{1}{\sqrt{2}}$,
as it is done for the MSE loss.
\textbf{Lower bounds:}
Computing the lower bound $l$ with
the $\lmcut$
heuristic.
When we use a Gaussian distribution,
$l$ is used to clip the heuristic prediction $\E[\rx]=\mu(s)$ to the interval $[l, \infty)$.
Ablation studies with $l=\hmax(s)$ \citep{bonet2001planning} and $l=\blind(s)$ are
included in the Appendix.
\textbf{Residual learning:}
Either using the model to directly predict $\mu(s)$
or to predict an offset $\Delta\mu(s)$ over a heuristic $h(s)$,
so that $\mu(s) = \Delta\mu(s) + h(s)$.
We use $h=\ff$ as our unbiased estimator of $h^*$, as proposed by \citet{yoon2008learning}.
In the Appendix, we conduct experiments with $\lmcut$ as the basis of the residual.

\paragraph{Training}
We trained each configuration with 5 different random seeds
on a training dataset that consists of
400 problem instances subsampled from the entire training problem set (456-1536 instances, depending on the domain).
Due to the nature of the dataset, these 400 problem instances can result in a different number of
data points depending on the length of the optimal plan of each instance.
We performed $4\times 10^4$ weight updates (training steps)
using \textit{AdamW} \citep{loshchilov2018fixing} with batch size 256,
weight decay $10^{-2}$ to avoid overfitting, gradient clip $0.1$,
learning rate of $10^{-2}$ for the linear regression and NLM, and $10^{-3}$ for HGN.
All models use the NLL loss for training, motivated by the theory,
but note that the NLL of $\N(\mu,\sigma=1/\sqrt{2})$
matches the MSE up to a constant, as previously noted.
For each model, we saved the weights that resulted in the best validation MSE metric during the training.
On a single NVIDIA Tesla V100,
each NLM training took $\approx 0.5$ hrs except in visitall ($\approx 2$ hrs).
HGN was much slower ($\approx 3$ hrs except $\approx 15$ hrs in blocksworld).
Linear models trained much faster (12-20 minutes).

\paragraph{Evaluation Scheme}
We first report two different metrics on the test set:
``MSE'' and ``MSE+clip''.
Here, MSE is the mean square error between $h^*(s_i)$ and $h(s_i)=\E[\rx]$,
i.e., $\frac{1}{N}\sum_{i=1}^{N} (h(s_i)-h^*(s_i))^2$,
for $i$-th state $s_i$ of $N$ states in the test dataset.
$\E[\rx]$ of $\T\N$ is given by \refeq{eq:tg-mean}
while $\E[\rx]$ of $\N$ is simply $\mu$.
``+clip'' variants are exclusive to $\N$ and they clip $\mu$ to $l$,
i.e., use $\max(\mu,l)$ in place of $\mu$ to compute the MSE.
We also obtained the MSE for $h=\ff$ and $h=\lmcut$.

We then evaluate the planning performance using the point estimate provided by each model as
a heuristic function to guide a search algorithm.
Since the learned heuristic is inadmissible,
we evaluate our heuristics in an agile search setting,
where Greedy Best-First Search \citep[GBFS]{bonet1999planning} is the standard algorithm.
We do not use \astar because
it does not guarantee finding the optimal (shortest) plan \citep{russell2010artificial}
with inadmissible heuristics
and it is slower than GBFS in the agile search
as it must explore all nodes below the current best $f=g+h$ value,
which is unnecessary for finding a satisficing solution.
In our experiments,
we evaluate search performance as
the combination of
the number of solved instances and the number of heuristic evaluations required to solve each instance,
with a limit of 10000 evaluations per problem.
We do not use runtime as our metric so that results are independent of the hardware and software configuration.
Additionally, we evaluated GBFS with the off-the-shelf $\ff$ heuristic as a baseline.
The planning component is based on Pyperplan \citep{pyperplan}.

\subsection{Heuristic Accuracy Evaluations}

\begin{table*}[tb]
\centering
 \adjustbox{width=\textwidth}{
 \begin{tabular}{lrll*{8}{c}}
 \toprule
 & & &
 & \multicolumn{2}{c}{learn/$\ff$} & \multicolumn{2}{c}{learn/none} & \multicolumn{2}{c}{fixed/$\ff$} & \multicolumn{2}{c}{fixed/none} \\
 \cmidrule(r){5-6} \cmidrule(r){7-8} \cmidrule(r){9-10} \cmidrule(r){11-12}
 {\small domain}
 & {\small metric}
 & $\ff$
 & $\lmcut$
 & $\N$ & $\T\N$ & $\N$ & $\T\N$ & $\N$ & $\T\N$ & $\N$ & $\T\N$  \\
\midrule {\small blocks} & MSE & 22.8 & 25.06 & .76$\pm$.1 & \textbf{.65$\pm$.1} & 3.26$\pm$.6 & \textbf{2.71$\pm$.4} & .83$\pm$.1 & \textbf{.66$\pm$.1} & 2.97$\pm$.9 & \textbf{2.44$\pm$.3} \\
 & +clip &  &  & .76$\pm$.2 &  & 2.91$\pm$.4 &  & .83$\pm$.2 &  & 2.74$\pm$.6 &  \\
 \midrule {\small ferry} & MSE & 9.77 & 11.10 & 3.73$\pm$.7 & \textbf{3.45$\pm$.8} & 141.05$\pm$29.4 & \textbf{8.63$\pm$2.7} & \textbf{2.98$\pm$1.4} & 3.85$\pm$.9 & 118.59$\pm$10.4 & \textbf{9.58$\pm$1.5} \\
 & +clip &  &  & 3.72$\pm$.6 &  & 10.44$\pm$28.4 &  & \textbf{2.98$\pm$1.1} &  & 10.50$\pm$9.6 &  \\
 \midrule {\small gripper} & MSE & 9.93 & 15.82 & 3.65$\pm$.9 & 3.70$\pm$.9 & 68.12$\pm$16.0 & \textbf{5.65$\pm$1.3} & 3.69$\pm$.9 & 3.72$\pm$.9 & 68.22$\pm$16.1 & \textbf{11.97$\pm$2.2} \\
 & +clip &  &  & 3.65$\pm$.7 &  & 13.37$\pm$15.2 &  & 3.69$\pm$.8 &  & 13.38$\pm$14.5 &  \\
 \midrule {\small visitall} & MSE & 13.9 & 36.4 & 7.67$\pm$.4 & \textbf{5.30$\pm$.6} & 25.31$\pm$7.9 & \textbf{9.70$\pm$1.6} & 6.49$\pm$.6 & 6.62$\pm$.9 & 21.71$\pm$2.6 & \textbf{14.11$\pm$1.0} \\
 & +clip &  &  & 7.60$\pm$.4 &  & 18.79$\pm$7.3 &  & \textbf{6.35$\pm$.6} &  & 16.38$\pm$2.3 &  \\
 \bottomrule
 \end{tabular}
 }
 \caption{
Test metrics for NLM (smaller the better). Each number represents the mean$\pm$std of 5 random seeds.
For each configuration, we performed $10^4$ training steps, saving the checkpoints
with the best validation MSE metric.
We tested several orthogonal configurations: 1) Learning $\sigma$ (\emph{learn}) or fixing it to $\frac{1}{\sqrt{2}}$ (\emph{fixed})
and 2) Using residual learning ($\ff$) or not (\emph{none}).
For each configuration, we compare the test MSE metric of the Gaussian
($\N$) and Truncated Gaussian ($\T\N$) models.
Rows labeled as \emph{+clip} denote a $\N$ model where $\mu$ is clipped above $\lmcut$.
For each configuration, the best average MSE among $\N$, $\N$+clip, and $\T\N$ is highlighted in \textbf{bold},
if the value gap to the second-best is larger than 0.1.
Results for linear regression and STRIPS-HGN models are provided in the Appendix.
 }
 \label{tbl:train}
\end{table*}

We focus on the results obtained by the NLM models,
as our conclusions from the Linear and the HGN models
(See Appendix)
were not substantially different.
\reftbl{tbl:train} shows the MSE metric of the heuristics obtained by different configurations
evaluated on the test instances (which are significantly larger than the training instances).
Compared to the models trained with the NLL loss of $\N$,
those trained with our proposed $\T\N$ loss often result in significantly more accurate heuristics.
For example,
in \pddl{ferry} and \pddl{gripper},
some $\N$ models completely fail to learn a useful heuristic, as shown by the large heuristic errors
(e.g., the base $\N$/fixed/none model on \pddl{ferry} obtains an MSE of 118.59).
In these situations, the clipping trick often reduces errors significantly
(e.g., the $\N+clip$/fixed/none model on the same domain obtains an MSE of 10.50).
However, this simply indicates that the $\N$ models are falling back to the $\lmcut$ heuristic
for those (many) predictions which are smaller than $\lmcut$.
This is why, even with clipping,
$\N$ models fail to match the accuracy of $\T\N$ models in many cases:
For example, the MSE of $\N$+clip/learn/none on \pddl{gripper} is 7.7 points larger than the one of $\T\N$/learn/none.
This confirms our hypothesis that admissible heuristics such as $\lmcut$ should be used as the lower bound of $\T\N$,
instead of simply to perform post-hoc clipping of heuristic predictions.

Additional detailed observations follow.
\textbf{First,}
$\T\N$ tends to converge faster during training, as shown in \refig{fig:loss-curve}.
\textbf{Second,}
residual learning often improves accuracy considerably,
thus proving to be an effective way of utilizing inadmissible heuristics.
\textbf{Third,}
we observed that the trained heuristics,
including those that use residual learning from $\ff$,
tend to be more accurate than $\ff$.
This rejects the hypothesis that residual learning is simply copying $\ff$ values.
\textbf{Fourth,}
learning $\sigma$ helps $\T\N$ exclusively.
For every $\N$ and $\T\N$ model,
\reftbl{tbl:train} contains
2 comparisons related to $\sigma$ (learn/none vs.\ fixed/none and learn/$\ff$ vs.\ fixed/$\ff$) across 4 domains,
resulting in a total of 8 comparisons.
Out of 8,
learning $\sigma$ degrades the MSE of $\N$ in 5 cases,
while it improves the MSE of $\T\N$ in 7 cases.
This happens because
$\sigma$ affects the expected value $\E[\rx]$ of $\T\N$ used as the heuristic prediction
but it does not for $\N$.
In other words,
$\T\N$ models requires both $\mu$ and $\sigma$ in order to achieve good heuristic accuracy.
This explains why $\T\N$/fixed/$\ff$ is not as competitive as $\N$/fixed/$\ff$:
fixed/$\ff$ is an ill-defined configuration for $\T\N$.

\subsection{Search Performance Evaluations}

\begin{table*}[tb]
\centering
\begin{tabular}{l|c|ccc|ccc}
\toprule
 & & \multicolumn{3}{c|}{learn/$\ff$ (proposed)} & \multicolumn{3}{c}{fixed/none (baseline)} \\
domain & $\ff$ & $\N$ & $\N$+clip & $\T\N$ & $\N$ & $\N$+clip & $\T\N$ \\
\midrule
\multicolumn{8}{c}{Ratio of solved instances under $10^4$ evaluations (higher the better)}\\
\midrule
blocks & .13 & .84$\pm$.19 & .85$\pm$.19 & \textbf{.88$\pm$.14} & \textbf{.79$\pm$.29} & .50$\pm$.35 & .55$\pm$.33 \\
ferry & .82 & .91$\pm$.19 & .91$\pm$.19 & \textbf{.98$\pm$.05} & .01$\pm$.01 & .57$\pm$.10 & \textbf{.58$\pm$.13} \\
gripper & .96 & \textbf{1} & \textbf{1} & \textbf{1} & 0 & .92$\pm$.12 & \textbf{1} \\
visitall & .86 & .97$\pm$.07 & \textbf{.98$\pm$.06} & \textbf{.98$\pm$.05} & .82$\pm$.33 & \textbf{1} & \textbf{1} \\
\midrule
\multicolumn{8}{c}{Average node evaluations (smaller the better)}\\
\midrule
blocks & 9309 & 2690$\pm$2128 & 2681$\pm$2121 & \textbf{2060$\pm$1607} & \textbf{4118$\pm$2663} & 6268$\pm$2675 & 5903$\pm$2685 \\
ferry & 5152 & 3216$\pm$1964 & 3117$\pm$1967 & \textbf{2477$\pm$1093} & 9933$\pm$92 & 6675$\pm$582 & \textbf{6475$\pm$725} \\
gripper & 3918 & 1642$\pm$139 & 1643$\pm$141 & \textbf{1637$\pm$492} & 10000$\pm$0 & 2941$\pm$1513 & \textbf{1709$\pm$658} \\
visitall & 3321 & 2156$\pm$1451 & 2148$\pm$1511 & \textbf{1683$\pm$1290} & 3384$\pm$3448 & \textbf{591$\pm$216} & 612$\pm$363 \\
\bottomrule
\end{tabular}
\caption{
Planning results on NLM weights saved according to the best validation MSE metric,
comparing the average$\pm$stdev of the ratio of solved instances under $10^4$ node evaluations
and the average number of evaluated nodes across problems.
The number of evaluated nodes is counted as $10^4$ on instances the planner failed to solve.
For each configuration (learn/$\ff$ or fixed/none), we highlight the best results in bold.
}
\label{tbl:plan}
\end{table*}

We compared the search performance of GBFS using heuristic functions obtained by the different models
as well as the \sota off-the-shelf $\ff$ heuristic.
We included
our proposed \emph{learn/$\ff$} configuration
and
the baseline \emph{fixed/none} configuration.
Results for \emph{learn/none} and \emph{fixed/$\ff$} can be found in the Appendix.
\reftbl{tbl:plan} shows the average$\pm$stdev of
the ratio of problem instances solved (i.e., coverage),
where a value of 1 means all instances are solved, and
the average number of node evaluations per problem over 5 seeds.
The second metric is introduced to differentiate between methods that solve most or all of the instances.

We observed that, with our proposed \emph{learn/$\ff$} configuration,
the learned heuristics significantly outperform the off-the-shelf $\ff$ heuristic.
Additionally, $\T\N$ outperforms $\N$ and $\N$+clip in every domain when both
the ratio of solved instances and number of node evaluations are considered
(the second metric is used to break ties in the first one).

Conversely, with the traditional but less ideal \emph{fixed/none} configuration,
several learned heuristics are surpassed by $\ff$ and, also,
$\T\N$ is outperformed by $\N$ or $\N$+clip in some cases.
These results align with those shown in \reftbl{tbl:train}.
Firstly, $\N$ models which do not use clipping sometimes learn dismal heuristics
(e.g., in \pddl{gripper}, $\N$/fixed/none fails to solve any instance).
Secondly, $\T\N$ models need to predict $\sigma$ (in addition to $\mu$)
in order to learn heuristics of good quality.

\section{Related Work}

Using admissible heuristics as lower bounds of a $\T\N$ distribution may appear trivial in the hindsight.
Existing work use $\T\N$ for machine learning
most often in the context of safety-aware planning,
where
the upper/lower bounds are \emph{arbitrary} constraints
imposed by the environment or by a domain expert.
For example,
\citet{murray2023column} uses $\T\N$ to model a Simple Temporal Network with Uncertainty (STNU)
which can model a distribution of time within a specific start time / deadline.
\citet{eisen2019learning} uses $\T\N$ to optimize wireless device allocations,
where the truncation encodes the range of signal power.
In robotics, $\T\N$ is often used
to limit the measurement uncertainty \citep{kamran2021minimizing}.

In contrast,
the admissible heuristics used as lower bounds in our work
are \emph{formal bounds automatically proved by} symbolic algorithms.
For example, $\lmcut$ is computed by
deriving a so-called landmark graph, and then
reducing the costs on the edges that constitute a cut of the graph.
To our knowledge, \emph{our work is the first to show that such
formally derived bounds for combinatorial tasks can be combined with $\T\N$.}

For instance,
in applications of machine learning to Operations Research problems (e.g., Vehicle Routing Problem, TSP),
existing work often tries to learn to solve them
without the help of heuristics \citep{nazari2018reinforcement}.
Although \citet{xin2021neurolkh} uses the optimal solution
obtained by a traditional optimal method (e.g. Concorde solver) for training and
combines it with existing admissible heuristics (LKH heuristic) during testing,
it does not use the heuristic for training.

In the context of heuristic learning in automated planning,
off-the-shelf heuristics have only been used as a training target \citep{shen2020learning},
or as a residual basis \citep{yoon2008learning}.

In Reinforcement Learning \citep{sutton2018reinforcement},
it is a common practice to accelerate the training through reward shaping,
which is theoretically equivalent to residual learning \citep{ng1999policy}.
An extension of reward shaping \citep{cheng2021hurl} utilizes hand-crafted heuristics.
\citet{gehring2021pddlrl} used $\ff$ to shape rewards for classical planning.
However, to our knowledge,
none has leveraged admissible heuristics as lower/upper bounds.
\citet{cheng2021hurl} also discussed the \emph{pessimistic} and \emph{admissible} heuristics
as desirable properties of RL and planning heuristics,
but their method does not explicitly use the upper/lower bound property for training.

\section{Conclusion and Future Work}

In this paper, we studied the problem of supervised heuristic learning under a statistical lens, focusing on how to effectively
utilize the information provided by admissible heuristics. Firstly, we provided some statistical background on heuristic
learning which was later leveraged to analyze the decisions made (sometimes unknowingly) in the literature. We explained how the
commonly used MSE loss implicitly models the heuristic to be learned as a Gaussian distribution.
Then, we argued that this heuristic should instead be modeled as a Truncated Gaussian,
where admissible heuristics are used as the lower bound of the distribution.
We conducted extensive experimentation, comparing the heuristics learned with our
truncated-based statistical model versus those learned by minimizing squared errors.
Results show that our proposed method improves convergence speed during training and
yields more accurate heuristics that result in better planning performance,
thus confirming that it is the correct approach for utilizing admissible bounds in heuristic learning.

Our findings serve to answer the three important questions we raised in the introduction:
\textbf{What should the model learn?}
To achieve super-symbolic benefits,
we should use expensive metrics such as $h^*$, not poly-time heuristics or sub-optimal plan costs.
\textbf{How should we train the model?}
We maximize the likelihood of the observed $h^*$ assuming a Truncated Gaussian distribution
lower bounded by an admissible heuristic.
\textbf{Why so?}
The \emph{principle of maximum entropy}:
the Truncated Gaussian distribution encodes our prior knowledge without any extra assumptions that may cause harmful bias.

In future work, we will extend our proposed method to other learning settings.
One interesting scenario is given by iterative search algorithms \citep{richter2010joy,richter2011lama},
where the cost of the best solution found so far could be used as the upper bound of a Truncated Gaussian.
Another avenue for future work is to explore the Reinforcement Learning setting
where a value function is learned instead of a heuristic,
extending the work on residual learning for RL \citep{gehring2021pddlrl}.

\section*{Acknowledgements}

This work has been partially funded by the Grant PID2022-142976OB-I00,
funded by MICIU/AEI/ 10.13039/501100011033 and by “ERDF/EU”,
as well as the Andalusian Regional predoctoral grant no. 21-111-PREDOC-0039 and by “ESF Investing in your future”.

\clearpage

\appendix

\section{Truncated Gaussian Max-Ent Proof}
\label{appendix:max_ent_proof}

The proof for $\T\N$ being the maximum-entropy (max-ent) distribution
with finite mean, finite variance and a support equal to $(l,u)$
can be found in \citep{dowson1973maximum},
as referenced in Section 3 of our main paper.
Its Theorem 1 states that every finite support max-ent distribution with particular first two moments $\mu_1$, $\mu_2$
takes a form $p(x)=\exp {-(\lambda_0+\lambda_1 x+\lambda_2 x^2)}$,
where $\lambda_0,\lambda_1,\lambda_2$ are chosen to satisfy
$\mu_n = \int x^n \exp {-(\lambda_0+\lambda_1 x+\lambda_2 x^2)} dx$, with $\mu_0=1$.
The first and the second moment are just synonyms of the mean $\mu$ and the variance $\sigma^2$ (other moments also have synonyms; $\mu_3$ is skewness).
Just below its Equation (5), they state that when $\lambda_2>0$, $p(x)$ corresponds to a TN.
We can indeed transform
our Equation (8) (definition of $\T\N$) into $p(x)$ above
by matching
$-\log p(x)= \lambda_0+\lambda_1 x+\lambda_2 x^2$ and our Equation (9):
$\lambda_0 = \frac{\mu^2}{2\sigma^2}+\log \sqrt{2\pi\sigma^2} + \log(\Phi(\frac{u-\mu}{\sigma})-\Phi(\frac{l-\mu}{\sigma}))$,
$\lambda_1 = -\frac{\mu}{\sigma^2}$,
and
$\lambda_2 = \frac{1}{2\sigma^2}$.

\section{Truncated Gaussian Implementation}
\label{appendix:trunc_gauss_implementation}

This Appendix explains several important implementation details of the Truncated Gaussian distribution used in our work.
The code for our Pytorch implementation can be found in Github\footnote{https://github.com/TheAeryan/stable-truncated-gaussian}
and is also provided as a PyPI package\footnote{https://pypi.org/project/stable-trunc-gaussian/}.

\subsection{Numerically Stable Formulas for Truncated Gaussian}

The Truncated Gaussian distribution is a four-parameter probability distribution defined as follows:
\begin{align*}
 \N(\rx | \mu,\sigma,l,u) &=
 \left\{
 \begin{array}{cc}
  \frac{1}{\sigma}
  \frac{\phi(\frac{\rx-\mu}{\sigma})}
       {\Phi(\frac{u-\mu}{\sigma})-\Phi(\frac{l-\mu}{\sigma})} & l \leq \rx \leq u\\
  0 & \text{otherwise.}
 \end{array}
 \right.\\
 \text{where}\
 \phi(\rx) &= \frac{1}{\sqrt{2\pi}} \exp \frac{\rx^2}{2},\\
 \Phi(\rx) &= \frac{1}{2} (1 + \erf(\rx)).
\end{align*}

In order to train and use a system that involves a Truncated Gaussian,
we need to compute several properties, such as its mean and the log-probability of some value $x$
under the distribution.
However,
the naive implementation of the formulas for calculating these quantities
is numerically unstable due to floating-point rounding errors,
especially when $\mu$ lies outside the interval $(l,u)$.
In this subsection, we briefly explain the source of instability and provide
numerically stable formulas for calculating these values.

Given a Truncated Gaussian distribution $\N(\rx\mid \mu,\sigma,l,u)$,
its mean $\E[\rx]$ is given by the following formula:
\begin{align*}
\E[\rx]
& =\mu+
  \frac{\phi(\alpha)-\phi(\beta)}
       {\Phi(\beta) -\Phi(\alpha)}\sigma, \quad \text{where}
\end{align*}
\[
\textstyle\alpha=\frac{l-\mu}{\sigma},\quad\beta=\frac{u-\mu}{\sigma}. \quad (\beta\geq\alpha)
\]
The expression $\frac{\phi(\alpha)-\phi(\beta)}{\Phi(\beta)-\Phi(\alpha)}$
should not be evaluated directly because
it involves subtractions between values that could be potentially very close to each other,
causing floating-point rounding errors.

We now describe a stable implementation of this formula introduced by \citet{Jorge2018}.
Let us define the following function:
\[
F_{1}\parens{x,y}=\frac{e^{-x^{2}}-e^{-y^{2}}}{\erf\parens{y}-\erf(x)}
\]
Then, we reformulate the mean as follows:
\begin{align*}
& \E[\rx]
=\mu + \sqrt{\frac{2}{\pi}}F_{1}\parens{\frac{\alpha}{\sqrt{2}},\frac{\beta}{\sqrt{2}}}\sigma.
\end{align*}

$F_1$ can be evaluated in a numerically stable manner by using the formulas below:
\begin{align*}
 & F_{1}(x,y) \\
 & =F_{1}(y,x), &  & \text{if }|x|>|y|\\
  & =P_{1}(x,y-x), &  & \text{if }|x-y|=|\epsilon|<10^{-7}\\
 & =\frac{1-\Delta}{\Delta\erfcx(-y)-\erfcx(-x)} &  & \text{if }x,y\le0\\
 & =\frac{1-\Delta}{\erfcx(x)-\Delta\erfcx(y)} &  & \text{if }x,y\ge0\\
 & =\frac{(1-\Delta)e^{-x^{2}}}{\erf(y)-\erf(x)} &  & \text{otherwise}.
\end{align*}
where
$\Delta=e^{x^{2}-y^{2}}$,
$\erfcx(x)=e^{x^2}\erfc(x)$ is a function that is commonly available in mathematical packages,
and
$P_1$ is a Taylor expansion of $F_{1}(x,x+\epsilon) = P_1(x,\epsilon)$ where $y=x+\epsilon$:
\begin{align*}
& P_{1}(x,\epsilon) =\sqrt{\pi}x+\frac{1}{2}\sqrt{\pi}\epsilon-\frac{1}{6}\sqrt{\pi}x\epsilon^{2}-\frac{1}{12}\sqrt{\pi}\epsilon^{3}+\\
&\frac{1}{90}\sqrt{\pi}x(x^{2}+1)\epsilon^{4}.
\end{align*}

Next, we provide a numerically stable method for computing the log-probability $\log \N(x\mid \mu,\sigma,l,u)$.
Let us assume $l \leq x \leq u$, since otherwise the probability is 0 (whose logarithm is $-\infty$).
The value is given by the following expression:
\begin{align*}
& \log\N(x\mid \mu,\sigma,l,u) = \log\Biggl(\frac{1}{\sigma}\frac{\phi(\xi)}{\Phi(\beta)-\Phi(\alpha)}\Biggr) =\\
& -\log \sigma-\log \sqrt{2\pi} -\frac{\xi^2}{2}-\log\bigl(\Phi(\beta)-\Phi(\alpha)\bigr), \\
&\text{where}\quad \xi=\frac{x-\mu}{\sigma}.
\end{align*}

Let $Z = \Phi(\beta)-\Phi(\alpha)$.
We obtain $\log(Z)$ from the stable formula for $\E[\rx]$.
When $\alpha, \beta \geq 0$,
\begin{align*}
 &\log(Z)= -\log \frac{\E[\rx]-\mu}{\sigma} -\log\sqrt{2\pi}  - \frac{\alpha^2}{2} + \\
 &\log\parens{1-e^{\frac{\alpha^2-\beta^2}{2}}}.
\end{align*}
When $\alpha, \beta \leq 0$,
\begin{align*}
 &\log(Z)= -\log \frac{\mu-\E[\rx]}{\sigma} -\log\sqrt{2\pi} - \frac{\beta^2}{2}  + \\
 &\log\parens{1-e^{\frac{\beta^2-\alpha^2}{2}}}.
\end{align*}
Otherwise,
\begin{align*}
 \log(Z)= -\log 2 + \log \left[ \erf\parens{\frac{\beta}{\sqrt{2}}}-\erf\parens{\frac{\alpha}{\sqrt{2}}} \right].
\end{align*}

\subsection{Truncated Gaussian with Missing Bounds}

A Truncated Gaussian distribution can be defined with either the lower $l$ or upper bound $u$ missing,
as $\N(\mu,\sigma,-\infty,u)$ or $\N(\mu,\sigma,l,\infty)$, respectively.
It can also be defined with no bounds at all as $\N(\mu,\sigma,-\infty,\infty)$,
in which case it is equivalent to an untruncated Gaussian $\N(\mu,\sigma)$.

In our implementation, we use $l=-1e5$ and $u=1e5$ as the parameters of a Truncated Gaussian
with no lower and/or upper bound, respectively.
We have observed that these values result indistinguishable from $l \to -\infty$ and $u \to \infty$
when calculating the mean $\E[\rx]$ and log-probability $\log p(x)$,
as long as $-1e5 \ll \mu \ll 1e5$, $\sigma \ll 1e5$ and $-1e5 < x < 1e5$
(since $p(x)=0$ for any $x$ outside the interval $[l,u]$).

\subsection{Truncated Gaussian with Open Bounds}

When defining a Truncated Gaussian distribution $\N(\mu,\sigma,l,u)$, we need to specify whether the bounds
$l,u$ are contained in the support of the distribution or not, i.e.,
whether the support is equal to $[l,u]$ (they are contained) or $(l,u)$ (they are \emph{not} contained).
When the support is $[l,u]$ we say that the Truncated Gaussian has \emph{closed bounds}
and that it has \emph{open bounds} otherwise.

Our first Truncated Gaussian implementation used closed bounds,
but we discovered that this decision would sometimes lead to learning issues
since the ML model would tend to output $\mu \ll 0$ (e.g., $\mu = -100$).
We believe the reason for that behavior is that a highly accurate lower bound $l$
 (e.g., $\lmcut$) can be sometimes equal to $h^*$
and the ML model is encouraged to maximize $\log p(l)=\log p(h^*)$.
In order to do so, it can simply output $\mu \ll 0$,
as the smaller (more negative) $\mu$ gets,
the higher $\log p(l)$ becomes.
Therefore, using closed bounds would often result in a learned heuristic equivalent to $l=\lmcut$,
as the mean of $\N(\mu,\sigma,l,u)$ is almost equal to $l$ when $\mu \ll l$.

For this reason, we switched to open bounds in our implementation.
To do so, we simply subtracted a small value $\epsilon=0.1$ from $l$,
obtaining a new distribution $\N(\mu,\sigma,l-\epsilon,u)$.
This made sure that $x$ was never equal to $l'=l-\epsilon$ when calculating $\log p(x=h^*)$,
which prevented the ML model from predicting $\mu \ll 0$.
Finally, in order to obtain a Truncated Gaussian where the upper bound $u$ is also \emph{open},
we can add $\epsilon$ to $u$,
which results in a new distribution $\N(\mu,\sigma,l-\epsilon,u+\epsilon)$.

\section{Parameter details}
\label{appendix:parameter_details}

\subsection{Model Hyperparameters}
\label{appendix:model_hyperparameters}

In this Appendix, we detail the hyperparameter values used for the different models: NLM, HGN, and linear regression.
In general, we did not perform extensive hyperparameter tuning for the different models.

For the NLM, we used a model with breadth 3 and depth 5,
where every inner layer outputs 8 predicates for each arity.
The multi-layer perceptrons used in the network employed sigmoid as their activation function
and contained no hidden layer.
AdamW \citep{loshchilov2018fixing} with weight decay 0.01 helped suppress overfitting.
(Weight decay causes an issue in Adam \citep{loshchilov2018fixing},
while AdamW fixes it.
Without the weight decay, Adam and AdamW are identical.)

For the HGN, we employed a \emph{hidden size} of 32 and 4 recursion steps.
We note that using more recursion steps did not improve the performance significantly
while being slower.
As mentioned in the main paper, the learning rate for the HGN is $1e^{-3}$,
which corresponds to the value used in \citet{shen2020learning}.

Finally, we report that
we initially tested an L2 weight decay penalty for the linear regression model
but removed it because it did not help it.

\subsection{Parameters of Instance Generators}
\label{appendix:instance_generator_params}

\begin{table}[tb]
 \centering
 \begin{tabular}{|c|p{20em}|}
  \toprule

  \multicolumn{2}{|c|}{blocksworld} \\
  \midrule
  train & $\textrm{seed}\in 1..38$, $\textrm{blocks}\in 5..16$ \\
  val & $\textrm{seed}\in 1..11$, $\textrm{blocks}\in 5..16$ \\
  test & $\textrm{seed}\in 1..11$, $\textrm{blocks}\in 11..22$ \\
  \midrule

  \multicolumn{2}{|c|}{ferry} \\
  \midrule
  train & $\textrm{seed}\in 1..16$, $\textrm{locations}\in 2..6$, $\textrm{cars}\in 2..6$ \\
  val & $\textrm{seed}\in 1..4$, $\textrm{locations}\in 2..6$, $\textrm{cars}\in 2..6$ \\
  test & $\textrm{seed}\in 1..16$, $\textrm{locations}\in \{10,15,20,25,30\}$, $\textrm{cars}\in \{10,15,20,25,30\}$ \\
  \midrule

  \multicolumn{2}{|c|}{gripper} \\
  \midrule
  train & $\textrm{seed}\in 1..80$, $\textrm{balls}\in \{2,4,6,8,10\}$ \\
  val & $\textrm{seed}\in 1..20$, $\textrm{balls}\in \{2,4,6,8,10\}$ \\
  test & $\textrm{seed}\in 1..20$, $\textrm{balls}\in \{20,40,60,80,100\}$ \\
  \midrule

  \multicolumn{2}{|c|}{visitall} \\
  \midrule
  train & $\textrm{seed}\in 1..70$, $x\in 3..5$, $x=y$, $\textrm{ratio}\in \{0.5, 1.0\}$ \\
  val & $\textrm{seed}\in 1..17$, $x\in 3..5$, $x=y$, $\textrm{ratio}\in \{0.5, 1.0\}$ \\
  test & $\textrm{seed}\in 1..17$, $x,y\in 5..7$, $\textrm{ratio}\in \{0.5, 1.0\}$ \\
  \bottomrule
 \end{tabular}
 \caption{List of instance generation parameters}
 \label{tab:generator-parameters}
\end{table}

We generated the problems used in our experiments with parameterized
generators \citep{fawcett-et-al-icaps2011wspal}.
\reftbl{tab:generator-parameters} shows the range of parameter values used for each generator.
Note that we modified \pddl{gripper} generator to select the initial states randomly from the entire state space,
unlike in the traditional instances whose initial state specifies that all balls are in the left room.

\onecolumn

\section{Full Experimental Results for NLMs}
\label{appendix:results_HGN}

In the main paper, we only show the NLM planning results for our proposed configuration (learn/$\ff$) and the baseline configuration (fixed/none) due to limited space.
\reftbl{tbl:plan_nlm} provides the full planning results for the NLM models, including the learn/none and fixed/$\ff$ configurations.
We also present the test accuracy results for consistency across all appendices.

Regarding the fixed/$\ff$ configuration,
we observe that $\T\N$ only obtains better planning results (when both the number of solved instances and evaluated nodes are compared) than $\N$ and $\N$+clip
on \pddl{visitall} and \pddl{gripper},
but not on \pddl{ferry} and \pddl{blocksworld}.
These results are not surprising.
After all, as commented in the main paper,
this configuration is problematic for $\T\N$
as $\sigma$ affects its heuristic prediction $\E[\rx]$ and, therefore,
it needs to learn $\sigma$ in order to achieve good performance.
This is also why it does not obtain better accuracy than the other approaches (See \reftbl{tbl:train_nlm}).

Regarding the learn/none configuration,
$\T\N$ is the best model overall:
It obtains better planning performance than the alternative approaches on \pddl{visitall} and \pddl{gripper},
comparable performance with $\N$+clip on \pddl{ferry}
and is outperformed on \pddl{blocksworld}.
Finally,
we note that the best configuration overall is the one we proposed:
learn/$\ff$.

\begin{table*}[h]
\centering
\begin{adjustbox}{width=\linewidth}
 \begin{tabular}{lrll*{8}{c}}
 \toprule
 & & &
 & \multicolumn{2}{c}{learn/$\ff$} & \multicolumn{2}{c}{learn/none} & \multicolumn{2}{c}{fixed/$\ff$} & \multicolumn{2}{c}{fixed/none} \\
 \cmidrule(r){5-6} \cmidrule(r){7-8} \cmidrule(r){9-10} \cmidrule(r){11-12}
 {\small domain}
 & {\small metric}
 & $\ff$
 & $\lmcut$
 & $\N$ & $\T\N$ & $\N$ & $\T\N$ & $\N$ & $\T\N$ & $\N$ & $\T\N$  \\
\midrule {\small blocks} & MSE & 22.8 & 25.06 & .76$\pm$.1 & \textbf{.65$\pm$.1} & 3.26$\pm$.6 & \textbf{2.71$\pm$.4} & .83$\pm$.1 & \textbf{.66$\pm$.1} & 2.97$\pm$.9 & \textbf{2.44$\pm$.3} \\
 & +clip &  &  & .76$\pm$.2 &  & 2.91$\pm$.4 &  & .83$\pm$.2 &  & 2.74$\pm$.6 &  \\
 \midrule {\small ferry} & MSE & 9.77 & 11.10 & 3.73$\pm$.7 & \textbf{3.45$\pm$.8} & 141.05$\pm$29.4 & \textbf{8.63$\pm$2.7} & \textbf{2.98$\pm$1.4} & 3.85$\pm$.9 & 118.59$\pm$10.4 & \textbf{9.58$\pm$1.5} \\
 & +clip &  &  & 3.72$\pm$.6 &  & 10.44$\pm$28.4 &  & \textbf{2.98$\pm$1.1} &  & 10.50$\pm$9.6 &  \\
 \midrule {\small gripper} & MSE & 9.93 & 15.82 & 3.65$\pm$.9 & 3.70$\pm$.9 & 68.12$\pm$16.0 & \textbf{5.65$\pm$1.3} & 3.69$\pm$.9 & 3.72$\pm$.9 & 68.22$\pm$16.1 & \textbf{11.97$\pm$2.2} \\
 & +clip &  &  & 3.65$\pm$.7 &  & 13.37$\pm$15.2 &  & 3.69$\pm$.8 &  & 13.38$\pm$14.5 &  \\
 \midrule {\small visitall} & MSE & 13.9 & 36.4 & 7.67$\pm$.4 & \textbf{5.30$\pm$.6} & 25.31$\pm$7.9 & \textbf{9.70$\pm$1.6} & 6.49$\pm$.6 & 6.62$\pm$.9 & 21.71$\pm$2.6 & \textbf{14.11$\pm$1.0} \\
 & +clip &  &  & 7.60$\pm$.4 &  & 18.79$\pm$7.3 &  & \textbf{6.35$\pm$.6} &  & 16.38$\pm$2.3 &  \\
 \bottomrule
 \end{tabular}
\end{adjustbox}
 \caption{
\textbf{Same as Table 1 of main paper (with updated std values).}
Test metrics for NLM (smaller the better). Each number represents the mean$\pm$std of 5 random seeds.
For each configuration, we performed $10^4$ training steps, saving the checkpoints
with the best validation MSE metric.
We tested several orthogonal configurations: 1) Learning $\sigma$ (\emph{learn}) or fixing it to $\frac{1}{\sqrt{2}}$ (\emph{fixed})
and 2) Using residual learning ($\ff$) or not (\emph{none}).
For each configuration, we compare the test MSE metric of the Gaussian
($\N$) and Truncated Gaussian ($\T\N$) models.
Rows labeled as \emph{+clip} denote a $\N$ model where $\mu$ is clipped above $\lmcut$.
For each configuration, the best average MSE among $\N$, $\N$+clip, and $\T\N$ is highlighted in \textbf{bold},
if the value gap to the second-best is larger than 0.1.
 }
 \label{tbl:train_nlm}
\end{table*}

\begin{table*}[h]
\centering
\begin{adjustbox}{width=\linewidth}
 \begin{tabular}{l|c|ccc|ccc|ccc|ccc}
 \toprule
 & & \multicolumn{3}{c|}{learn/$\ff$} & \multicolumn{3}{c}{learn/none} & \multicolumn{3}{c|}{fixed/$\ff$} & \multicolumn{3}{c}{fixed/none} \\
 domain & $\ff$ & $\N$ & $\N$+clip & $\T\N$ & $\N$ & $\N$+clip & $\T\N$ & $\N$ & $\N$+clip & $\T\N$ & $\N$ & $\N$+clip & $\T\N$ \\
 \midrule
 \multicolumn{14}{c}{Ratio of solved instances under $10^4$ evaluations (higher the better)}\\
 \midrule
blocks & .13 & .84$\pm$.18 & .85$\pm$.18 & \textbf{.88$\pm$.17} & \textbf{.85$\pm$.37} & .57$\pm$.36 & .51$\pm$.37 & \textbf{.90$\pm$.13} & .89$\pm$.13 & .87$\pm$.12 & \textbf{.79$\pm$.34} & .50$\pm$.34 & .55$\pm$.35 \\
ferry & .82 & .91$\pm$.00 & .91$\pm$.01 & \textbf{.98$\pm$.00} & .01$\pm$.05 & \textbf{.60$\pm$.12} & .59$\pm$.15 & \textbf{.99$\pm$.00} & \textbf{.99$\pm$.00} & .94$\pm$.01 & .01$\pm$.06 & .57$\pm$.16 & \textbf{.58$\pm$.08} \\
gripper & .96 & \textbf{1} & \textbf{1} & \textbf{1} & 0 & .75$\pm$.34 & \textbf{1} & \textbf{1} & \textbf{1} & \textbf{1} & .00$\pm$.00 & .92$\pm$.26 & \textbf{1} \\
visitall & .86 & .97$\pm$.07 & \textbf{.98$\pm$.08} & \textbf{.98$\pm$.06} & .79$\pm$.16 & \textbf{1} & \textbf{1} & .98$\pm$.03 & .99$\pm$.03 & \textbf{1} & .82$\pm$.14 & \textbf{1} & \textbf{1} \\
 \midrule
 \multicolumn{14}{c}{Average node evaluations (smaller the better)}\\
 \midrule
blocks & 9309 & 2690$\pm$2111 & 2681$\pm$2115 & \textbf{2060$\pm$1823} & \textbf{3225$\pm$3401} & 5754$\pm$2825 & 6492$\pm$2695 & \textbf{1791$\pm$1678} & 1809$\pm$1674 & 2171$\pm$1489 & \textbf{4118$\pm$3005} & 6268$\pm$2649 & 5903$\pm$2835 \\
ferry & 5152 & 3216$\pm$532 & 3117$\pm$557 & \textbf{2477$\pm$369} & 9952$\pm$299 & \textbf{6751$\pm$610} & 6780$\pm$952 & 2218$\pm$255 & \textbf{2197$\pm$255} & 3471$\pm$482 & 9933$\pm$340 & 6675$\pm$1019 & \textbf{6475$\pm$593} \\
gripper & 3918 & 1642$\pm$198 & 1643$\pm$210 & \textbf{1637$\pm$244} & 10000$\pm$0 & 4313$\pm$1685 & \textbf{1480$\pm$1484} & 1631$\pm$65 & 1635$\pm$60 & \textbf{1301$\pm$48} & 10000$\pm$0 & 2941$\pm$1360 & \textbf{1709$\pm$428} \\
visitall & 3321 & 2156$\pm$1458 & 2148$\pm$1491 & \textbf{1683$\pm$1269} & 3860$\pm$2393 & 818$\pm$353 & \textbf{385$\pm$616} & 1437$\pm$1114 & 1355$\pm$1119 & \textbf{1142$\pm$929} & 3384$\pm$2035 & \textbf{591$\pm$151} & 612$\pm$160 \\
 \bottomrule
 \end{tabular}
\end{adjustbox}
\caption{
Planning results for NLM model.
For each model, we use the weights that resulted in the best validation MSE loss during training.
Table columns and rows have the same meaning as in Table 2 of the main paper.
For each configuration, the best metric among $\N$, $\N\text{+clip}$ and $\T\N$ is highlighted in \textbf{bold}.
}
\label{tbl:plan_nlm}
\end{table*}

\clearpage

\section{Full Experimental Results for HGNs}
\label{appendix:results_HGN}

In this Appendix, we provide the results of the experiments conducted on the STRIPS-HGN models.
HGN models are trained with learning rate $1e^{-3}$.
The accuracy metrics obtained by the HGN model are shown in \reftbl{tbl:train_hgn},
whereas \reftbl{tbl:plan_hgn} shows its planning results.

The training of HGN models was highly unstable even with a reduced learning rate $1e^{-3}$
(compared to $1e^{-2}$ in other models)
and the results vary significantly across domains.
For example, in \pddl{ferry}, HGN models generally failed to learn effective heuristics,
as shown in \reftbl{tbl:train_hgn} where the MSE exceeds $10^7$.
In contrast, in \pddl{visitall}, it generally achieved better accuracy than NLMs.

One potential reason for the failure in \pddl{ferry} is the limited expressivity of HGN.
Unlike NLM, HGN is designed to only receive the delete-relaxed information about the problem instance,
which may harm its ability to learn a heuristic for the original instance.
Another potential reason for its weak performance is the training length:
Due to the reduced learning rate,
it may need more training steps in order to converge.

Focusing on the planning results,
\reftbl{tbl:plan_hgn} shows improvements of HGN-based heuristics over $\ff$ in \pddl{blocksworld} and \pddl{visitall}.
\pddl{ferry} results are poor due to the converge failure.

Finally,
$\T\N$ tends to improve the accuracy and planning performance over $\N$ and $\N$+clip with equivalent configurations.

\begin{table*}[h]
\centering
\begin{adjustbox}{width=\linewidth}
 \begin{tabular}{lrll*{8}{c}}
 \toprule
 & & &
 & \multicolumn{2}{c}{learn/$\ff$} & \multicolumn{2}{c}{learn/none} & \multicolumn{2}{c}{fixed/$\ff$} & \multicolumn{2}{c}{fixed/none} \\
 \cmidrule(r){5-6} \cmidrule(r){7-8} \cmidrule(r){9-10} \cmidrule(r){11-12}
 {\small domain}
 & {\small metric}
 & $\ff$
 & $\lmcut$
 & $\N$ & $\T\N$ & $\N$ & $\T\N$ & $\N$ & $\T\N$ & $\N$ & $\T\N$  \\
\midrule {\small blocks}    & MSE   & 22.8 & 25.06 & 2.3$\pm$2.6         & \textbf{1.5$\pm$.4} & \textbf{2.3$\pm$.8} & 3.9$\pm$2.4           & \textbf{2.6$\pm$2.3} & 6.5$\pm$10.8    & 5.2$\pm$2.6         & \textbf{1.8$\pm$.6} \\
                            & +clip &      &       & 2.3$\pm$2.6         &                     & \textbf{2.3$\pm$.8} &                       & \textbf{2.6$\pm$2.3} &                 & 3.7$\pm$1.8         &                     \\
 \midrule {\small ferry}    & MSE   & 9.77 & 11.10 & (8.9$\pm$1.8)$e^{8}$     & \textbf{(2.3$\pm$4.9)$e^{7}$}     & \textbf{(4.9$\pm$7.4)$e^{7}$}     & (5.2$\pm$9.2)$e^{7}$       & (3.0$\pm$3.8)$e^{5}$      & \textbf{(1.7$\pm$2.1)$e^{7}$} & (4.2$\pm$6.0)$e^{6}$     & \textbf{(2.7$\pm$4.8)$e^{7}$}     \\
                            & +clip &      &       & (8.9$\pm$1.8)$e^{8}$     &                     & \textbf{(4.9$\pm$7.4)$e^{7}$}     &                       & (2.4$\pm$2.8)$e^{5}$      &                 & (4.2$\pm$6.0)$e^{6}$     &                     \\
 \midrule {\small gripper}  & MSE   & 9.93 & 15.82 & 1.1$\pm$.4          & 1.5$\pm$.6          & 59.2$\pm$91.3       & \textbf{7.5$\pm$12.6} & 3.7$\pm$2.5          & 4.4$\pm$6.0     & 9.4$\pm$12.8        & 3.0$\pm$2.7         \\
                            & +clip &      &       & \textbf{1.0$\pm$.4} &                     & 58.5$\pm$90.4       &                       & \textbf{3.5$\pm$2.6} &                 & \textbf{2.5$\pm$.8} &                     \\
 \midrule {\small visitall} & MSE   & 13.9 & 36.4  & 4.0$\pm$.4          & \textbf{3.6$\pm$.5} & \textbf{.3$\pm$.0}  & \textbf{.3$\pm$.0}    & 4.3$\pm$.5           & 4.1$\pm$.4      & \textbf{.3$\pm$.0}  & \textbf{.3$\pm$.1}  \\
                            & +clip &      &       & 3.8$\pm$.4          &                     & \textbf{.3$\pm$.0}  &                       & \textbf{4.0$\pm$.5}  &                 & \textbf{.3$\pm$.0}  &                     \\
 \bottomrule
 \end{tabular}
\end{adjustbox}
 \caption{
Test metrics for HGN model.
Table columns and rows have the same meaning as in Table 1 of the main paper.
For each configuration, the best metric among $\N$, $\N+clip$, $\T\N$ is highlighted in \textbf{bold}.
 }
 \label{tbl:train_hgn}
\end{table*}

\begin{table*}[h]
\centering
\begin{adjustbox}{width=\linewidth}
 \begin{tabular}{l|c|ccc|ccc|ccc|ccc}
 \toprule
 & & \multicolumn{3}{c|}{learn/$\ff$} & \multicolumn{3}{c}{learn/none} & \multicolumn{3}{c|}{fixed/$\ff$} & \multicolumn{3}{c}{fixed/none} \\
 domain & $\ff$ & $\N$ & $\N$+clip & $\T\N$ & $\N$ & $\N$+clip & $\T\N$ & $\N$ & $\N$+clip & $\T\N$ & $\N$ & $\N$+clip & $\T\N$ \\
 \midrule
 \multicolumn{14}{c}{Ratio of solved instances under $10^4$ evaluations (higher the better)}\\
 \midrule
blocks & .13 & .70$\pm$.30 & \textbf{.72$\pm$.29} & .48$\pm$.26 & .39$\pm$.22 & .42$\pm$.25 & \textbf{.57$\pm$.35} & .57$\pm$.28 & .62$\pm$.32 & \textbf{.66$\pm$.29} & .38$\pm$.29 & .39$\pm$.34 & \textbf{.62$\pm$.35} \\
ferry & .82 & .01$\pm$.02 & .01$\pm$.02 & \textbf{.02$\pm$.06} & .00$\pm$.00 & .00$\pm$.00 & \textbf{.00$\pm$.02} & .02$\pm$.04 & \textbf{.06$\pm$.12} & .02$\pm$.04 & .01$\pm$.02 & \textbf{.02$\pm$.03} & .01$\pm$.03 \\
gripper & .96 & .36$\pm$.14 & .37$\pm$.15 & \textbf{.39$\pm$.13} & .24$\pm$.20 & .24$\pm$.20 & \textbf{.27$\pm$.11} & .25$\pm$.14 & .27$\pm$.18 & \textbf{.35$\pm$.25} & .19$\pm$.06 & .38$\pm$.28 & \textbf{.63$\pm$.40} \\
visitall & .86 & \textbf{.99$\pm$.03} & .97$\pm$.04 & .97$\pm$.04 & \textbf{1} & \textbf{1} & \textbf{1} & .98$\pm$.03 & .99$\pm$.03 & \textbf{1} & \textbf{1} & \textbf{1} & \textbf{1} \\
 \midrule
 \multicolumn{14}{c}{Average node evaluations (smaller the better)}\\
 \midrule
blocks & 9309 & 3984$\pm$2675 & \textbf{3906$\pm$2649} & 5844$\pm$2088 & 6636$\pm$1859 & 6456$\pm$1969 & \textbf{5186$\pm$2915} & 5235$\pm$2262 & 5007$\pm$2439 & \textbf{4335$\pm$2548} & 6649$\pm$2554 & 6576$\pm$2812 & \textbf{4581$\pm$3096} \\
ferry & 5152 & 9916$\pm$132 & 9915$\pm$134 & \textbf{9834$\pm$424} & 10000$\pm$0 & 10000$\pm$0 & \textbf{9968$\pm$105} & 9897$\pm$248 & \textbf{9659$\pm$644} & 9834$\pm$258 & 9924$\pm$169 & \textbf{9856$\pm$250} & 9941$\pm$175 \\
gripper & 3918 & 7078$\pm$971 & 7008$\pm$1058 & \textbf{6949$\pm$1020} & 8050$\pm$1490 & 8048$\pm$1492 & \textbf{7952$\pm$564} & 8074$\pm$801 & 7917$\pm$1138 & \textbf{7271$\pm$1852} & 8264$\pm$474 & 7300$\pm$1633 & \textbf{4997$\pm$3206} \\
visitall & 3321 & 1512$\pm$1192 & 1555$\pm$1149 & \textbf{1472$\pm$1182} & 204$\pm$144 & 155$\pm$24 & \textbf{146$\pm$12} & 1158$\pm$955 & 1063$\pm$858 & \textbf{875$\pm$761} & 166$\pm$41 & \textbf{152$\pm$21} & 153$\pm$15 \\
 \bottomrule
 \end{tabular}
\end{adjustbox}
\caption{
Planning results for HGN model.
Table columns and rows have the same meaning as in Table 2 of the main paper.
For each configuration, the best metric among $\N$, $\N\text{+clip}$ and $\T\N$ is highlighted in \textbf{bold}.
}
\label{tbl:plan_hgn}
\end{table*}

\clearpage

\section{Full Experimental Results for Linear Regression (LR)}
\label{appendix:results_linear_regression}

\reftbl{tbl:train_rr} shows both the MSE and the NLL metrics obtained by the LR models on the test problems.
To evaluate the NLL,
we used the best validation NLL loss checkpoints.
We show the NLL here because it is the holistic way to compare the training and test results,
since all the models (including $\N$) are trained by minimizing the NLL loss, and not the MSE.
We used the MSE as our accuracy metric, and not the NLL, in the main paper for two main reasons.
First, MSE is easier to understand and interpret than NLL.
For example, if a model obtains an MSE of $2$ this means that its learned heuristic deviates from $h^*$
by $2$ units on average, whereas an NLL of $2$ has no straightforward interpretation.
Second, the MSE and NLL tend to correlate.
However, this latter condition is violated in the LR models,
so we decided to show both the NLL and MSE metrics.
We observe that $\T\N$ outperforms $\N$ according to the NLL metric.

We now detail additional observations.
The LR models are remarkably accurate in gripper compared to the NLM models despite their simplicity.
Residual learning provided no benefit to LR models
because they already receive the $\ff$ heuristic as one of their inputs,
so using $\ff$ again as the basis for the residual does not provide any additional information.
We also observe that learning $\sigma$ tends to improve the NLL.

\reftbl{tbl:plan_rr} shows the search performance of the LR models
using best validation MSE checkpoints.
$\T\N$ and $\N$ models showed comparable performance,
except in fixed $\sigma$ configurations where $\T\N$ models suffered.
Due to the simple model architecture (a single linear layer),
all models with different random seeds converged to the same search behavior,
despite the different weight initialization.
Indeed, the standard deviations in \reftbl{tbl:train_rr} also tend to be significantly smaller compared to other models.

\begin{table*}[h]
\centering
\begin{adjustbox}{width=0.75\linewidth}
 \begin{tabular}{lrll*{8}{c}}
 \toprule
 & & &
 & \multicolumn{2}{c}{learn/$\ff$} & \multicolumn{2}{c}{learn/none} & \multicolumn{2}{c}{fixed/$\ff$} & \multicolumn{2}{c}{fixed/none} \\
 \cmidrule(r){5-6} \cmidrule(r){7-8} \cmidrule(r){9-10} \cmidrule(r){11-12}
 {\small domain}
 & {\small metric}
 & $\ff$
 & $\lmcut$
 & $\N$ & $\T\N$ & $\N$ & $\T\N$ & $\N$ & $\T\N$ & $\N$ & $\T\N$  \\
 \midrule {\small blocks} & MSE & 22.8 & 25.06 & \textbf{6.5$\pm$.1} & 6.7$\pm$.4 & \textbf{6.3$\pm$.0} & 6.8$\pm$.4 & 7.0$\pm$.1 & 7.8$\pm$.2 & 6.9$\pm$.1 & 7.8$\pm$.1 \\
 & +clip &  &  & \textbf{6.5$\pm$.1} &  & \textbf{6.3$\pm$.0} &  & \textbf{6.9$\pm$.1} &  & \textbf{6.8$\pm$.1} &  \\
 \midrule {\small ferry} & MSE & 9.77 & 11.10 & 1.3$\pm$.1 & 3.1$\pm$.6 & \textbf{1.1$\pm$.3} & 3.0$\pm$.6 & \textbf{1.0$\pm$.3} & 8.5$\pm$.8 & \textbf{1.0$\pm$.2} & 8.5$\pm$.6 \\
 & +clip &  &  & \textbf{1.2$\pm$.1} &  & \textbf{1.1$\pm$.3} &  & \textbf{1.0$\pm$.3} &  & \textbf{1.0$\pm$.2} &  \\
 \midrule {\small gripper} & MSE & 9.93 & 15.82 & \textbf{.5$\pm$.2} & .7$\pm$.3 & \textbf{.6$\pm$.2} & .7$\pm$.3 & \textbf{.5$\pm$.2} & 1.2$\pm$.3 & \textbf{.5$\pm$.3} & 1.1$\pm$.4 \\
 & +clip &  &  & \textbf{.5$\pm$.2} &  & \textbf{.6$\pm$.2} &  & \textbf{.5$\pm$.2} &  & \textbf{.5$\pm$.3} &  \\
 \midrule {\small visitall} & MSE & 13.9 & 36.4 & 6.6$\pm$.1 & \textbf{5.0$\pm$.2} & 2.0$\pm$.0 & 2.9$\pm$.0 & 6.5$\pm$.1 & \textbf{6.4$\pm$.1} & 2.0$\pm$.0 & 2.3$\pm$.1 \\
 & +clip &  &  & 6.6$\pm$.1 &  & \textbf{1.9$\pm$.0} &  & 6.4$\pm$.1 &  & \textbf{1.9$\pm$.0} &  \\
 \midrule
 \midrule {\small blocks} & NLL & 22.8 & 25.06 & 2.2$\pm$.0 & \textbf{1.6$\pm$.0} & 2.1$\pm$.0 & \textbf{1.6$\pm$.0} & 3.0$\pm$.0 & \textbf{2.8$\pm$.0} & 3.0$\pm$.0 & \textbf{2.8$\pm$.0} \\
 & +clip &  &  & 2.2$\pm$.0 &  & 2.1$\pm$.0 &  & 3.0$\pm$.0 &  & 3.0$\pm$.0 &  \\
 \midrule {\small ferry} & NLL & 9.77 & 11.10 & 1.6$\pm$.1 & \textbf{1.4$\pm$.1} & 1.5$\pm$.1 & \textbf{1.4$\pm$.1} & \textbf{1.5$\pm$.1} & 3.1$\pm$.2 & \textbf{1.5$\pm$.1} & 3.1$\pm$.2 \\
 & +clip &  &  & 1.6$\pm$.1 &  & 1.5$\pm$.1 &  & \textbf{1.5$\pm$.1} &  & \textbf{1.5$\pm$.1} &  \\
 \midrule {\small gripper} & NLL & 9.93 & 15.82 & 1.0$\pm$.1 & \textbf{.9$\pm$.2} & 1.4$\pm$.2 & \textbf{.9$\pm$.2} & 1.4$\pm$.1 & \textbf{1.2$\pm$.1} & 1.4$\pm$.1 & \textbf{1.1$\pm$.1} \\
 & +clip &  &  & 1.0$\pm$.1 &  & 1.4$\pm$.2 &  & 1.4$\pm$.1 &  & 1.4$\pm$.1 &  \\
 \midrule {\small visitall} & NLL & 13.9 & 36.4 & 2.1$\pm$.0 & \textbf{1.8$\pm$.0} & \textbf{1.5$\pm$.0} & \textbf{1.5$\pm$.1} & 2.9$\pm$.0 & \textbf{2.7$\pm$.0} & 1.8$\pm$.0 & \textbf{1.6$\pm$.0} \\
 & +clip &  &  & 2.1$\pm$.0 &  & \textbf{1.5$\pm$.0} &  & 2.9$\pm$.0 &  & 1.7$\pm$.0 &  \\
 \bottomrule
 \end{tabular}
\end{adjustbox}
 \caption{
Test metrics for Linear Regression (LR) model.
Table columns and rows have the same meaning as in Table 1 of the main paper.
For each configuration, the best metric among $\N$, $\N+clip$, $\T\N$ is highlighted in \textbf{bold}.
 }
 \label{tbl:train_rr}
\end{table*}

\begin{table*}[h]
\centering
\begin{adjustbox}{width=\linewidth}
 \begin{tabular}{l|c|ccc|ccc|ccc|ccc}
 \toprule
 & & \multicolumn{3}{c|}{learn/$\ff$} & \multicolumn{3}{c}{learn/none} & \multicolumn{3}{c|}{fixed/$\ff$} & \multicolumn{3}{c}{fixed/none} \\
 domain & $\ff$ & $\N$ & $\N$+clip & $\T\N$ & $\N$ & $\N$+clip & $\T\N$ & $\N$ & $\N$+clip & $\T\N$ & $\N$ & $\N$+clip & $\T\N$ \\
 \midrule
 \multicolumn{14}{c}{Ratio of solved instances under $10^4$ evaluations (higher the better)}\\
 \midrule
blocks & .13 & \textbf{.20$\pm$.00} & .18$\pm$.00 & .17$\pm$.00 & \textbf{.21$\pm$.00} & .20$\pm$.00 & .19$\pm$.00 & \textbf{.20$\pm$.00} & .19$\pm$.00 & .19$\pm$.00 & \textbf{.21$\pm$.00} & .17$\pm$.00 & .19$\pm$.00 \\
ferry & .82 & \textbf{1} & \textbf{1} & \textbf{1} & \textbf{1} & \textbf{1} & \textbf{1} & \textbf{1} & \textbf{1} & .45$\pm$.00 & \textbf{1} & \textbf{1} & .45$\pm$.00 \\
gripper & .96 & \textbf{1} & \textbf{1} & \textbf{1} & \textbf{1} & \textbf{1} & \textbf{1} & \textbf{1} & \textbf{1} & .76$\pm$.00 & \textbf{1} & \textbf{1} & .70$\pm$.00 \\
visitall & .86 & .91$\pm$.00 & .91$\pm$.00 & \textbf{.96$\pm$.00} & \textbf{.92$\pm$.00} & .91$\pm$.00 & \textbf{.92$\pm$.00} & \textbf{.91$\pm$.00} & \textbf{.91$\pm$.00} & \textbf{.91$\pm$.00} & .90$\pm$.00 & \textbf{.91$\pm$.00} & \textbf{.91$\pm$.00} \\
 \midrule
 \multicolumn{14}{c}{Average node evaluations (smaller the better)}\\
 \midrule
blocks & 9309 & \textbf{8770$\pm$0} & 8920$\pm$0 & 8989$\pm$0 & \textbf{8658$\pm$0} & 8795$\pm$0 & 8951$\pm$0 & \textbf{8844$\pm$0} & 8849$\pm$0 & 8984$\pm$0 & \textbf{8816$\pm$0} & 8992$\pm$0 & 8944$\pm$0 \\
ferry & 5152 & 1891$\pm$0 & \textbf{1886$\pm$0} & 1944$\pm$0 & 1894$\pm$0 & \textbf{1891$\pm$0} & 2025$\pm$0 & \textbf{1862$\pm$0} & 1900$\pm$0 & 7652$\pm$0 & 1877$\pm$0 & \textbf{1860$\pm$0} & 7630$\pm$0 \\
gripper & 3918 & \textbf{964$\pm$0} & 968$\pm$0 & 973$\pm$0 & 972$\pm$0 & \textbf{970$\pm$0} & 971$\pm$0 & \textbf{966$\pm$0} & \textbf{966$\pm$0} & 4587$\pm$0 & 969$\pm$0 & \textbf{964$\pm$0} & 4775$\pm$0 \\
visitall & 3321 & 2829$\pm$0 & 3048$\pm$0 & \textbf{2035$\pm$0} & 3040$\pm$0 & 2952$\pm$0 & \textbf{2783$\pm$0} & 3132$\pm$0 & \textbf{2772$\pm$0} & 2859$\pm$0 & 3079$\pm$0 & 2947$\pm$0 & \textbf{2853$\pm$0} \\
 \bottomrule
 \end{tabular}
\end{adjustbox}
\caption{
Planning results for Linear Regression (LR) model.
For each model, we use the weights that resulted in the best validation MSE loss during training.
Table columns and rows have the same meaning as in Table 2 of the main paper.
For each configuration, the best metric among $\N$, $\N\text{+clip}$ and $\T\N$ is highlighted in \textbf{bold}.
}
\label{tbl:plan_rr}
\end{table*}

\clearpage
\section{Comparison with Ranking-based Approaches, with a Focus on Instance-Size Generalization}
\label{appendix:results_crestein}

We compare our approach with a recent ranking-based approach \citep{chrestien2023optimize}.
While cost-to-go approaches learn to estimate the distance from some particular state to the goal,
ranking-based heuristic learning \citep{garrett2016learning} tries to optimize the search performance
by directly learning the node ordering.

We emphasize that \emph{learning-to-rank methods deviate significantly from the cost-to-go learning literature}
(which our work belongs to), both empirically and theoretically,
as they pursue a different goal (correct node ordering vs correct cost-to-go estimation)
and are significantly outside the scope of our work.
Our paper aims at addresssing the lack of formal understanding in the cost-to-go learning literature
and, in section 3, paragraph 8, we explicitly state that our goal is to learn the cost-to-go $p^*(x|s)$.

Being able to estimate the cost-to-go is useful in many practical applications in its own right,
e.g., knowing the estimated travel time to destination in a mobile navigation app (e.g., Google Maps),
as opposed to just
knowing which direction to go at every intersection (ranking the directions).
Applying our theoretical intuition from $\mathcal{TN}$ to the learning-to-rank approach
is an interesting direction for future work.

\paragraph{Implementation}
We evaluated a Julia-based implementation published by the authors (\url{https://github.com/aicenter/Optimize-Planning-Heuristics-to-Rank}).
We also minimally modified SymbolicPlanner.jl library so that we can limit the number of node evaluations to 10000,
a condition being used throughout our paper.

Their implementation uses HGNs, thus it is best to compare it with our HGN results.
However, there are still minor differences:
First, they used their own re-implementation of HGN in Julia language
which may have unknown differences from
the Python implementation written by the original authors \citep{shen2020learning}.
Indeed, they used so-called \emph{hyper-graph reduplication} technique \citep{sourek2018lifted} to speed up the training.
In their paper, they implemented a residual connection technique for the HGN,
which we do not use as preliminary experiments showed that it degrades performance.
They also performed a grid-search to find the best hyperparameters
(recursion steps, dense layer size, use of residual connections).
Instead, we used the same set of hyperparameters as our HGN experiments.

\paragraph{Training}

\citet{chrestien2023optimize} proposed two ranking-based losses $L^*$ and $L^{\text{GBFS}}$.
$L^*$ and $L^{\text{GBFS}}$ are designed for and perform the best with \astar and GBFS.
Since we focus our evaluation on GBFS,
we only show the planning performance of heuristics trained by $L^{\text{GBFS}}$.
We call this configuration the \emph{HGN-$L^{\text{GBFS}}$}.

We trained their system on our training instances and evaluated its planning performance on our test instances,
which are significantly larger than the training ones. 
Conversely, \citet{chrestien2023optimize} did not evaluate instance-size generalization,
as the authors trained and tested on problems of roughly the same difficulty.

\paragraph{Results}

\reftbl{tbl:plan_crestein} shows the planning performance of HGN-$L^{\text{GBFS}}$ and cost-to-go models.
In \pddl{blocksworld}, HGN-$L^{\text{GBFS}}$ was outperformed by cost-to-go models.
In \pddl{ferry},
HGN-$L^{\text{GBFS}}$ is outperformed by the $\ff$ baseline in the number of problems solved
despite evaluating slightly fewer nodes overall, which may indicate overfitting
(i.e., when HGN-$L^{\text{GBFS}}$ finds a solution, it does so quickly).
Since this domain is also where cost-to-go models do not perform well,
it supports our hypothesis in \refsec{appendix:results_HGN}
that HGN cannot effectively process \pddl{ferry} in an instance-size generalized manner.
In \pddl{gripper},
their system failed to solve any instance due to
high branching factor and a potential memory leakage.
We tried to run them on our cluster with up to 512GB memory per process to no avail.
Finally,
in \pddl{visitall},
HGN-$L^{\text{GBFS}}$ completely failed to solve any instance
while HGN-$\N$/+clip/$\T\N$ solved most instances.

From these results, we conclude that our cost-to-go approach
overall outperforms the ranking-based approach proposed in \citet{chrestien2023optimize}.
Nonetheless, we reemphasize that
both methods follow different paradigms
and, thus, should be complementary in principle.
In the future,
we will study how to develop a method that both
1) optimizes search performance by directly learning the correct node orderings
and 2) utilizes the information provided by admissible heuristics to improve learning.

\begin{table*}[h]
\centering
\begin{adjustbox}{width=0.6\linewidth}
\begin{tabular}{l|c|ccc|c|}
\toprule
& & \multicolumn{3}{c|}{learn/$\ff$} & \citep{chrestien2023optimize}\\
domain & $\ff$ & $\N$ & $\N$+clip & $\T\N$ & $L^{GBFS}$ \\
\midrule
\multicolumn{6}{c}{Ratio of solved instances under $10^4$ evaluations (higher the better)}\\
\midrule
blocks   & .13 & .70$\pm$.30          & \textbf{.72$\pm$.29} & .48$\pm$.26          & .24 \\
ferry    & \textbf{.82} & .01$\pm$.02          & .01$\pm$.02          & {.02$\pm$.06} & .61 \\
gripper  & \textbf{.96} & .36$\pm$.14          & .37$\pm$.15          & {.39$\pm$.13} & --  \\
visitall & .86 & \textbf{.99$\pm$.03} & .97$\pm$.04          & .97$\pm$.04          & 0   \\
\midrule
\multicolumn{6}{c}{Average node evaluations (smaller the better)}\\
\midrule
blocks   & 9309 & 3984$\pm$2675 & \textbf{3906$\pm$2649} & 5844$\pm$2088          & 8282.6 \\
ferry    & 5152 & 9916$\pm$132  & 9915$\pm$134           & {9834$\pm$424}  & \textbf{5103.5} \\
gripper  & \textbf{3918} & 7078$\pm$971  & 7008$\pm$1058          & {6949$\pm$1020} & --     \\
visitall & 3321 & 1512$\pm$1192 & 1555$\pm$1149          & \textbf{1472$\pm$1182} & 10000  \\
\bottomrule
\end{tabular}
\end{adjustbox}
\caption{
Planning results of ranking-based HGN model ($L^{GBFS}$)
compared with our distributional HGN models ($\N$, $\N\text{+clip}$ and $\T\N$).
Table columns and rows have the same meaning as in Table 2 of the main paper.
For each configuration, the best metric among $\ff$, $\N$, $\N\text{+clip}$, $\T\N$, and $L^{GBFS}$ is highlighted in \textbf{bold}.
}
\label{tbl:plan_crestein}
\end{table*}

\clearpage
\section{Experimental Results with Different Lower Bounds}
\label{appendix:other_l_experiments}

\reftbls{tbl:train_other_l}{tbl:plan_other_l} show the ablation study
of the NLM models using $\hmax$ and $\blind$ instead of $\lmcut$ as the lower bound $l$,
with our proposed (and best) learn/$\ff$ configuration.

We observe that the $\N$+clip models obtain almost identical MSE
regardless of the heuristic used for the clipping.
This makes sense:
As commented in the main paper,
$\N$+clip is identical to $\N$ except for 
those cases where the model does a \textit{really bad} prediction, i.e.,
predicts a cost-to-go that is very far off from the target $h^*$.
If the ML model has been trained correctly,
this should seldom occur,
meaning that $\N$+clip will be equivalent to $\N$ and, thus,
no matter the heuristic ($\lmcut$, $\hmax$ or $\blind$) used for the clipping,
the performance will be the same (on average).

Regarding the $\T\N$ models,
we observe that they obtain comparable test MSE on \pddl{blocksworld}, \pddl{ferry} and \pddl{gripper}.
However, on \pddl{visitall},
the $l=\lmcut$ configuration clearly outperforms the other two:
$l=\lmcut$ obtains an MSE of 5.3, $l=\hmax$ a value of 7.01 and $l=\blind$ of 7.15.
We believe this may be due to the difficulty of \pddl{visitall},
as this is the domain for which the different models obtain highest MSE.
Due to its high difficulty,
the use of a more informative bound $l$, i.e., $\lmcut$ instead of $\hmax$ or $\blind$, may make a big difference for the $\T\N$ models,
thus resulting in more accurate predictions.

\begin{table*}[h]
\centering
 \begin{tabular}{lrll*{10}{c}}
 \toprule
 & & &
 & \multicolumn{2}{c}{learn/$\ff$ ($l=\lmcut$)} & \multicolumn{2}{c}{learn/$\ff$ ($l=\hmax$)} & \multicolumn{2}{c}{learn/$\ff$ ($l=\blind$)} \\
 \cmidrule(r){5-6} \cmidrule(r){7-8} \cmidrule(r){9-10}
 {\small domain}
 & {\small metric}
 & $\ff$
 & $\lmcut$
 & $\N$ & $\T\N$ & $\N$ & $\T\N$ & $\N$ & $\T\N$ \\
\midrule {\small blocks} & MSE & 22.8 & 25.06 & .76$\pm$.1 & \textbf{.65$\pm$.1} & \textbf{.76$\pm$.1} & \textbf{.76$\pm$.2} & .76$\pm$.1 & \textbf{.66$\pm$.1} \\
 & +clip &  &  & .76$\pm$.2 & & \textbf{.76$\pm$.1} &  & .76$\pm$.1 &  \\
 \midrule {\small ferry} & MSE & 9.77 & 11.10 & 3.73$\pm$.7 & \textbf{3.45$\pm$.8} & 3.98$\pm$.9 & \textbf{3.13$\pm$.9} & 3.73$\pm$.8 & \textbf{3.21$\pm$.9} \\
 & +clip &  &  & 3.72$\pm$.6 & & 3.98$\pm$.9 &  & 3.73$\pm$.8 &  \\
 \midrule {\small gripper} & MSE & 9.93 & 15.82 & \textbf{3.65$\pm$.9} & 3.70$\pm$.9 & \textbf{3.66$\pm$.9} & 3.75$\pm$1.0 & \textbf{3.65$\pm$.9} & 3.69$\pm$1.1 \\
 & +clip &  &  & \textbf{3.65$\pm$.7} & & \textbf{3.66$\pm$.9} &  & \textbf{3.65$\pm$.9} &  \\
 \midrule {\small visitall} & MSE & 13.9 & 36.4 & 7.67$\pm$.4 & \textbf{5.30$\pm$.6} & 7.58$\pm$.5 & \textbf{7.01$\pm$.3} & 7.55$\pm$.4 & \textbf{7.15$\pm$.5} \\
 & +clip &  &  & 7.60$\pm$.4 & & 7.55$\pm$.5 &  & 7.55$\pm$.4 &  \\
 \bottomrule
 \end{tabular}
 \caption{
Test metrics with $l=\lmcut$, $l=\hmax$ and $l=\blind$, using the \emph{learn/$\ff$} configuration.
Table columns and rows have the same meaning as in Table 1 of the main paper.
For each configuration, the best metric among $\N$, $\N+clip$, $\T\N$ is highlighted in \textbf{bold}.
Results for $\N$ models may vary across different $l$ due to the use of different seeds.
 }
 \label{tbl:train_other_l}
\end{table*}

\begin{table*}[h]
\centering
\begin{adjustbox}{width=\linewidth}
\begin{tabular}{l|c|ccc|ccc|ccc}
\toprule
 & & \multicolumn{3}{c|}{learn/$\ff$ ($l=\lmcut$)} & \multicolumn{3}{c|}{learn/$\ff$ ($l=\hmax$)} & \multicolumn{3}{c}{learn/$\ff$ ($l=\blind$)} \\
domain & $\ff$ & $\N$ & $\N$+clip & $\T\N$ & $\N$ & $\N$+clip & $\T\N$ & $\N$ & $\N$+clip & $\T\N$ \\
\midrule
\multicolumn{8}{c}{Ratio of solved instances under $10^4$ evaluations (higher the better)}\\
\midrule
blocks & .13 & .84$\pm$.19 & .85$\pm$.19 & \textbf{.88$\pm$.14} & .82$\pm$.18 & .82$\pm$.18 & \textbf{.85$\pm$.16} & .83$\pm$.17 & .83$\pm$.17 & \textbf{.89$\pm$.13} \\
ferry & .82 & .91$\pm$.19 & .91$\pm$.19 & \textbf{.98$\pm$.05} & .97$\pm$.06 & .97$\pm$.06 & \textbf{.98$\pm$.05} & .90$\pm$.20 & .90$\pm$.20 & \textbf{.98$\pm$.02} \\
gripper & .96 & \textbf{1} & \textbf{1} & \textbf{1} & \textbf{1} & \textbf{1} & \textbf{1} & \textbf{1} & \textbf{1} & \textbf{1} \\
visitall & .86 & .97$\pm$.07 & \textbf{.98$\pm$.06} & \textbf{.98$\pm$.05} & .98$\pm$.05 & \textbf{.99$\pm$.03} & .98$\pm$.04 & .97$\pm$.05 & .98$\pm$.04 & \textbf{.99$\pm$.02} \\
\midrule
\multicolumn{8}{c}{Average node evaluations (smaller the better)}\\
\midrule
blocks & 9309 & 2690$\pm$2128 & 2681$\pm$2121 & \textbf{2060$\pm$1607} & 2871$\pm$2235 & 2863$\pm$2227 & \textbf{2480$\pm$1783} & 2816$\pm$2065 & 2805$\pm$2084 & \textbf{1987$\pm$1400} \\
ferry & 5152 & 3216$\pm$1964 & 3117$\pm$1967 & \textbf{2477$\pm$1093} & 2802$\pm$1115 & 2772$\pm$1080 & \textbf{2414$\pm$1051} & 3277$\pm$1989 & 3291$\pm$1998 & \textbf{1907$\pm$479} \\
gripper & 3918 & 1642$\pm$139 & 1643$\pm$141 & \textbf{1637$\pm$492} & 1891$\pm$553 & 1890$\pm$556 & \textbf{1763$\pm$361} & 1889$\pm$544 & 1896$\pm$560 & \textbf{1495$\pm$69} \\
visitall & 3321 & 2156$\pm$1451 & 2148$\pm$1511 & \textbf{1683$\pm$1290} & 2007$\pm$1301 & \textbf{1777$\pm$1154} & 1894$\pm$1313 & 1899$\pm$1292 & 1864$\pm$1330 & \textbf{1755$\pm$1132} \\
\bottomrule
\end{tabular}
\end{adjustbox}
\caption{
Planning results with $l=\lmcut$, $l=\hmax$ and $l=\blind$, using the \emph{learn/$\ff$} configuration.
Table columns and rows have the same meaning as in Table 2 of the main paper.
For each configuration, the best metric among $\N$, $\N\text{+clip}$ and $\T\N$ is highlighted in \textbf{bold}.
Results for $\N$ models may vary across different $l$ due to the use of different seeds.
}
\label{tbl:plan_other_l}
\end{table*}

\clearpage
\section{Experimental Results with Different Residuals}
\label{appendix:other_res_experiments}

\reftbls{tbl:train_other_res}{tbl:plan_other_res} show the ablation study
of the NLM models using $\lmcut$ as the residual base.
While $\lmcut$ is not theoretically ideal (compared to $\ff$)
as it is a biased estimator that is
always smaller than the target $h^*$,
in practice it also worked well as the residual base for heuristic learning.

We observe that, for the $\T\N$ model,
the $\ff$ residual results in better test MSE than the $\lmcut$ one
in every domain except \pddl{visitall}.
Conversely, we observe that $\lmcut$ residual works better for the $\N$ model,
as it obtains better MSE in every domain except \pddl{gripper}.
Therefore, it seems that the best heuristic to use as a base for the residual ($\ff$ vs $\lmcut$)
depends on the chosen model ($\T\N$ or $\N$).

Regardless of the heuristic employed,
the residual-based configurations (learn/$\ff$ and learn/$\lmcut$)
outperform learn/none,
thus showing the benefits of using residual learning.

\begin{table*}[h]
\centering
 \begin{tabular}{lrll*{6}{c}}
 \toprule
 & & &
 & \multicolumn{2}{c}{learn/$\ff$} & \multicolumn{2}{c}{learn/$\lmcut$} & \multicolumn{2}{c}{learn/none} \\
 \cmidrule(r){5-6} \cmidrule(r){7-8} \cmidrule(r){9-10}
 {\small domain}
 & {\small metric}
 & $\ff$
 & $\lmcut$
 & $\N$ & $\T\N$ & $\N$ & $\T\N$ & $\N$ & $\T\N$  \\
\midrule {\small blocks} & MSE & 22.8 & 25.06 & .76$\pm$.1 & \textbf{.65$\pm$.1} & .72$\pm$.1 & \textbf{.66$\pm$.1} & 3.26$\pm$.6 & \textbf{2.71$\pm$.4} \\
 & +clip &  &  & .76$\pm$.1 &  & .72$\pm$.1 &  & 2.91$\pm$.4 &  \\
 \midrule {\small ferry} & MSE & 9.77 & 11.10 & 3.73$\pm$.8 & \textbf{3.45$\pm$.8} & \textbf{3.23$\pm$1.1} & 3.58$\pm$1.3 & 141.05$\pm$29.6 & \textbf{8.63$\pm$2.7} \\
 & +clip &  &  & 3.72$\pm$.8 &  & \textbf{3.23$\pm$1.1} &  & 10.44$\pm$1.8 &  \\
 \midrule {\small gripper} & MSE & 9.93 & 15.82 & \textbf{3.65$\pm$.9} & 3.70$\pm$1.1 & 3.94$\pm$1.0 & \textbf{3.88$\pm$1.1} & 68.12$\pm$15.3 & \textbf{5.65$\pm$1.1} \\
 & +clip &  &  & \textbf{3.65$\pm$.9} &  & 3.94$\pm$1.0 &  & 13.37$\pm$2.2 &  \\
 \midrule {\small visitall} & MSE & 13.9 & 36.4 & 7.67$\pm$.4 & \textbf{5.30$\pm$.6} & 4.40$\pm$.8 & \textbf{4.03$\pm$.5} & 25.31$\pm$7.9 & \textbf{9.70$\pm$1.6} \\
 & +clip &  &  & 7.60$\pm$.4 &  & 4.40$\pm$.8 &  & 18.79$\pm$4.2 &  \\
 \bottomrule
 \end{tabular}
 \caption{
Test metrics using the \emph{learn/$\ff$}, \emph{learn/$\lmcut$}, \emph{learn/none} configurations.
Table columns and rows have the same meaning as in Table 1 of the main paper.
For each configuration, the best metric among $\N$, $\N+clip$, $\T\N$ is highlighted in \textbf{bold}.
 }
 \label{tbl:train_other_res}
\end{table*}

\begin{table*}[h]
\centering
\begin{adjustbox}{width=\linewidth}
 \begin{tabular}{l|c|ccc|ccc|ccc}
 \toprule
 & & \multicolumn{3}{c}{learn/$\ff$} & \multicolumn{3}{c}{learn/$\lmcut$} & \multicolumn{3}{c}{learn/none} \\
 domain & $\ff$ & $\N$ & $\N$+clip & $\T\N$ & $\N$ & $\N$+clip & $\T\N$ & $\N$ & $\N$+clip & $\T\N$ \\
 \midrule
 \multicolumn{10}{c}{Ratio of solved instances under $10^4$ evaluations (higher the better)}\\
 \midrule
 blocks & .13 & .84$\pm$.18 & .85$\pm$.18 & \textbf{.88$\pm$.15} & \textbf{.88$\pm$.14} & \textbf{.88$\pm$.15} & .85$\pm$.19 & \textbf{.85$\pm$.22} & .57$\pm$.36 & .51$\pm$.37 \\
 ferry & .82 & .91$\pm$.19 & .91$\pm$.19 & \textbf{.98$\pm$.04} & .64$\pm$.11 & .66$\pm$.15 & \textbf{.68$\pm$.12} & .01$\pm$.01 & \textbf{.60$\pm$.12} & .59$\pm$.13 \\
 gripper & .96 & \textbf{1} & \textbf{1} & \textbf{1} & \textbf{1} & \textbf{1} & \textbf{1} & .00$\pm$.00 & .75$\pm$.42 & \textbf{1} \\
 visitall & .86 & .97$\pm$.06 & \textbf{.98$\pm$.05} & \textbf{.98$\pm$.04} & \textbf{1} & \textbf{1} & \textbf{1} & .79$\pm$.34 & \textbf{1} & \textbf{1} \\
 \midrule
 \multicolumn{10}{c}{Average node evaluations (smaller the better)}\\
 \midrule
 blocks & 9309 & 2690$\pm$2193 & 2681$\pm$2192 & \textbf{2060$\pm$1673} & 2461$\pm$1747 & \textbf{2390$\pm$1757} & 2735$\pm$2106 & \textbf{3225$\pm$2446} & 5754$\pm$2723 & 6492$\pm$2693 \\
 ferry & 5152 & 3216$\pm$2033 & 3117$\pm$1991 & \textbf{2477$\pm$1209} & 6107$\pm$664 & 6088$\pm$739 & \textbf{5944$\pm$680} & 9952$\pm$70 & \textbf{6751$\pm$740} & 6780$\pm$863 \\
 gripper & 3918 & 1642$\pm$212 & 1643$\pm$218 & \textbf{1637$\pm$390} & 1392$\pm$349 & \textbf{1390$\pm$341} & 1477$\pm$482 & 10000$\pm$0 & 4313$\pm$3475 & \textbf{1480$\pm$378} \\
 visitall & 3321 & 2156$\pm$1431 & 2148$\pm$1550 & \textbf{1683$\pm$1245} & 508$\pm$205 & 531$\pm$232 & \textbf{483$\pm$162} & 3860$\pm$3249 & 818$\pm$562 & \textbf{385$\pm$112} \\
 \bottomrule
 \end{tabular}
\end{adjustbox}
\caption{
Planning results using the \emph{learn/$\ff$}, \emph{learn/$\lmcut$}, \emph{learn/none} configurations.
Table columns and rows have the same meaning as in Table 2 of the main paper.
For each configuration, the best metric among $\N$, $\N\text{+clip}$ and $\T\N$ is highlighted in \textbf{bold}.
}
\label{tbl:plan_other_res}
\end{table*}

\twocolumn
\section{Domain descriptions}
\label{appendix:domain_descriptions}

In this Appendix, we provide detailed descriptions and PDDL encodings for the four planning domains employed in our experiments: \pddl{blocksworld-4ops}, \pddl{ferry}, \pddl{gripper} and \pddl{visitall}.

\subsection{Blocksworld-4ops}

\pddl{Blocksworld} is one of the oldest domains in the planning literature.
It represents a table with a collection of blocks that can be stacked on top of each other.
The goal in this domain is to rearrange the blocks to achieve a specific configuration,
starting from some initial block arrangement.
The initial and goal configurations are randomized.
Blocks can be placed on top of another block or on the table,
and every block can never have more than a single block on top of it.
The arm/crane used to move the blocks around can only carry a single block at a time.
Listing \ref{lst:bw_domain} contains the PDDL description of this domain.

\begin{lstlisting}[
  float=!htb,
  caption={PDDL domain for \pddl{blocksworld-4ops}.},
  label={lst:bw_domain},
  language=PDDL]
(define (domain blocksworld-4ops)
  (:requirements :strips)
(:predicates (clear ?x)
             (on-table ?x)
             (arm-empty)
             (holding ?x)
             (on ?x ?y))

(:action pickup
  :parameters (?ob)
  :precondition (and (clear ?ob) (on-table ?ob) (arm-empty))
  :effect (and (holding ?ob) (not (clear ?ob)) (not (on-table ?ob)) (not (arm-empty))))

(:action putdown
  :parameters  (?ob)
  :precondition (holding ?ob)
  :effect (and (clear ?ob) (arm-empty) (on-table ?ob)
               (not (holding ?ob))))

(:action stack
  :parameters  (?ob ?underob)
  :precondition (and (clear ?underob) (holding ?ob))
  :effect (and (arm-empty) (clear ?ob) (on ?ob ?underob)
               (not (clear ?underob)) (not (holding ?ob))))

(:action unstack
  :parameters  (?ob ?underob)
  :precondition (and (on ?ob ?underob) (clear ?ob) (arm-empty))
  :effect (and (holding ?ob) (clear ?underob)
               (not (on ?ob ?underob)) (not (clear ?ob)) (not (arm-empty)))))
\end{lstlisting}

\subsection{Ferry}

\pddl{Ferry} is another classical domain.
The goal is to use a ferry to transport a series of cars from their start to their final locations,
which are randomized.
Each location is connected to every other location and the ferry can only carry one car at a time.
Listing \ref{lst:ferry_domain} contains the PDDL description of this domain.

\begin{lstlisting}[
  float=!htb,
  caption={PDDL domain for \pddl{ferry}.},
  label={lst:ferry_domain},
  language=PDDL]
(define (domain ferry)
  (:predicates (not-eq ?x ?y)
   (car ?c)
   (location ?l)
   (at-ferry ?l)
   (at ?c ?l)
   (empty-ferry)
   (on ?c))

  (:action sail
      :parameters  (?from ?to)
      :precondition (and (not-eq ?from ?to)
                         (location ?from) (location ?to) (at-ferry ?from))
      :effect (and  (at-ferry ?to)
        (not (at-ferry ?from))))

  (:action board
      :parameters (?car ?loc)
      :precondition  (and  (car ?car) (location ?loc)
         (at ?car ?loc) (at-ferry ?loc) (empty-ferry))
      :effect (and (on ?car)
       (not (at ?car ?loc))
       (not (empty-ferry))))

  (:action debark
      :parameters  (?car  ?loc)
      :precondition  (and  (car ?car) (location ?loc)
         (on ?car) (at-ferry ?loc))
      :effect (and (at ?car ?loc)
       (empty-ferry)
       (not (on ?car)))))
\end{lstlisting}

\subsection{Gripper}

In the \pddl{gripper} domain, a robot with two gripper hands must transport a series of balls
from one room to another.
Unlike in traditional \pddl{gripper} instances, the initial and goal location of each ball is randomized.
Each gripper hand can only carry one ball at a time.
Listing \ref{lst:gripper_domain} contains the PDDL description of this domain.

\begin{lstlisting}[
  float=!htb,
  caption={PDDL domain for \pddl{gripper}.},
  label={lst:gripper_domain},
  language=PDDL]
(define (domain gripper)
  (:predicates (room ?r)
   (ball ?b)
   (gripper ?g)
   (at-robby ?r)
   (at ?b ?r)
   (free ?g)
   (carry ?o ?g))

  (:action move
      :parameters  (?from ?to)
      :precondition (and  (room ?from) (room ?to) (at-robby ?from))
      :effect (and  (at-robby ?to)
        (not (at-robby ?from))))

  (:action pick
      :parameters (?obj ?room ?gripper)
      :precondition  (and  (ball ?obj) (room ?room) (gripper ?gripper)
         (at ?obj ?room) (at-robby ?room) (free ?gripper))
      :effect (and (carry ?obj ?gripper)
       (not (at ?obj ?room))
       (not (free ?gripper))))

  (:action drop
      :parameters  (?obj  ?room ?gripper)
      :precondition  (and  (ball ?obj) (room ?room) (gripper ?gripper)
         (carry ?obj ?gripper) (at-robby ?room))
      :effect (and (at ?obj ?room)
       (free ?gripper)
       (not (carry ?obj ?gripper)))))
\end{lstlisting}

\subsection{Visitall}

This deceptively simple domain describes a square \textit{NxN} grid in which a robot can move in the four directions
(up, down, right or left).
The goal is for the robot to visit several cells of the grid.
The initial robot location, the number of cells to visit and their positions in the grid
are all randomized.
Listing \ref{lst:visitall_domain} contains the PDDL description of this domain.

\begin{lstlisting}[
  float=!htb,
  caption={PDDL domain for \pddl{visitall}.},
  label={lst:visitall_domain},
  language=PDDL]
(define (domain visit-all)
  (:requirements :typing)
  (:types        place - object)
  (:predicates (connected ?x ?y - place)
         (at-robot ?x - place)
         (visited ?x - place)
  )
  
  (:action move
  :parameters (?curpos ?nextpos - place)
  :precondition (and (at-robot ?curpos) (connected ?curpos ?nextpos))
  :effect (and (at-robot ?nextpos) (not (at-robot ?curpos)) (visited ?nextpos))
  ))
\end{lstlisting}

\end{document}